\documentclass[10pt,journal,compsoc]{IEEEtran}
\usepackage{amsmath,amssymb,amsfonts}
\usepackage{amsmath, amsthm, amssymb}
\usepackage{graphicx}
\usepackage{textcomp}
\usepackage{xcolor}
\usepackage{subcaption}
\usepackage{float}
\usepackage{url}
\usepackage[inline]{enumitem}
\usepackage[symbol]{footmisc}
\usepackage[ruled,vlined]{algorithm2e}
\usepackage{mathtools}
\usepackage{cleveref}
\usepackage{textcomp}
\newtheorem{definition}{Definition}

\usepackage{multirow}

\ifCLASSOPTIONcompsoc
  \usepackage[nocompress]{cite}
\else
  \usepackage{cite}
\fi

\ifCLASSINFOpdf
\else
\fi

\begin{document}
\title{Locally Differentially Private Distributed Deep Learning via Knowledge Distillation}
\author{Di~Zhuang,
        Mingchen~Li
        and~J.~Morris~Chang,~\IEEEmembership{Senior~Member,~IEEE}
        % <-this % stops a space
\IEEEcompsocitemizethanks{\IEEEcompsocthanksitem Di Zhuang is with Snap Inc., 2772 Donald Douglas Loop N, Santa Monica, CA 90405. E-mail: zhuangdi1990@gmail.com.
\IEEEcompsocthanksitem Mingchen Li and J. Morris Chang are with the Department of Electrical Engineering, University of South Florida, Tampa, FL 33620. E-mail: \{mingchenli, chang5\}@usf.edu.
\IEEEcompsocthanksitem This work was done by Di Zhuang when he was at the University of South Florida.}% <-this % stops an unwanted space
%\thanks{Manuscript received May 7, 2019.}
}

\IEEEtitleabstractindextext{%
\begin{abstract}
Deep learning often requires a large amount of data. In real-world applications, e.g., healthcare applications, the data collected by a single organization (e.g., hospital) is often limited, and the majority of massive and diverse data is often segregated across multiple organizations. As such, it motivates the researchers to conduct distributed deep learning, where the data user would like to build DL models using the data segregated across multiple different data owners. However, this could lead to severe privacy concerns due to the sensitive nature of the data, thus the data owners would be hesitant and reluctant to participate. We propose LDP-DL, a privacy-preserving distributed deep learning framework via local differential privacy and knowledge distillation, where each data owner learns a teacher model using its own (local) private dataset, and the data user learns a student model to mimic the output of the ensemble of the teacher models. In the experimental evaluation, a comprehensive comparison has been made among our proposed approach (i.e., LDP-DL), DP-SGD, PATE and DP-FL, using three popular deep learning benchmark datasets (i.e., CIFAR10, MNIST and FashionMNIST). The experimental results show that LDP-DL consistently outperforms the other competitors in terms of privacy budget and model accuracy.
\end{abstract}

}

\maketitle

\IEEEdisplaynontitleabstractindextext

% \fi
\IEEEpeerreviewmaketitle

\IEEEraisesectionheading{\section{Introduction}\label{sec:ldp_introduction}}
\IEEEPARstart{D}{eep} learning (DL) has been shown to achieve extraordinary results in a variety of real-world applications, such as skin lesion analysis \cite{perez2019solo}, active authentication \cite{wu2016cost}, facial recognition \cite{nguyen2019autogan, zhuang2017fripal}, botnet detection \cite{zhuang2017peerhunter, zhuang2018enhanced} and community detection \cite{zhuang2019dynamo}. In traditional DL environment, training data is held by a single organization in a centralized fashion, that executes the DL algorithms. In general, a DL model would be more accurate and robust if it has been trained with more massive and more diverse data. However, in certain real-world applications, e.g., healthcare applications, the data collected by a single organization (e.g., hospital) is often limited, and the majority of massive and diverse data is often segregated across multiple organizations. As such, it motivates the researchers to conduct DL in a distributed fashion, where the data user (e.g., researcher/organization) would like to build DL models using the data segregated across multiple different data owners (e.g., organizations). 
However, the data owners would be hesitant and reluctant to participate in the data user's distributed deep learning, if the data user's protocol cannot resolve the data owners' important privacy concerns of their data. 
For instance, it has been shown that the private information could be inferred during the learning process \cite{8835245}, and the membership of certain training data could be traced back from the resulting trained model \cite{shokri2017membership}. Hence, it is imperative to design an effective privacy-preserving distributed deep learning approach. 

% \begin{table*}[!t]
% \footnotesize
% \captionsetup{font=footnotesize}
% %\renewcommand{\arraystretch}{1.3}
% \caption{Description of Facebook Social Network [Notations: $|V|$ ($|E|$): $\#$ of unique vertices (edges); $\mathbb{E}[|\triangle V|]$ ($\mathbb{E}[|\triangle E|]$): avg. $\#$ of vertices (edges) changed per network snapshots; $\#$ of snapshots: total number of consecutive network snapshots; time-interval: period of time between two consecutive network snapshots; time-span: total time spanning of each network dataset].}
% \label{table:EvolvingNetworks}
% \centering
% \begin{tabular}{c|c|c|c|c|c|c|c|c|c}
% \hline
%  \bfseries networks & $\mathbf{|V|}$ & $\mathbf{\mathbb{E}[|\triangle V|]}$ & \bfseries vertex-type & $\mathbf{|E|}$  & $\mathbf{\mathbb{E}[|\triangle E|]}$ & \bfseries edge-type & \bfseries $\#$ of snapshots & \bfseries time-interval & \bfseries time-span\\
%  \hline
%  \bfseries Facebook  & 59,302 & 12,765 & user  & 592,406  & 20,943 & friendship & 28 & 1 month & 28 months \\
% \hline
% \end{tabular}
% \end{table*}

Designing an effective and efficient privacy-preserving distributed deep learning approach is highly challenging. 
To date, a few approaches \cite{papernot2016semi, chase2017private, geyer2017differentially} have been proposed for privacy-preserving (distributed) deep learning. 
Papernot et al. \cite{papernot2016semi} proposes PATE, a ``teacher-student'' paradigm for privacy-preserving deep learning, where each data owner learns a teacher model using its own (local) private dataset, and the data user aims to learn a student model using the unlabelled public data (but no direct access to the data owners' private data) to mimic the output of the ensemble of the teacher models, i.e., the student learns to make predictions that is the same as the most number of teachers. To ensure privacy, PATE \cite{papernot2016semi} assumes a trusted aggregator to provide a differentially private query interface, where the data user could query the ensemble of the teacher models (from the data owners) using the unlabelled public data to obtain the labels for the training of the student model. However, a fully trusted aggregator barely exists in most of the real-world distributed deep learning scenarios. 
Chase et al. \cite{chase2017private} proposes a private collaborative neural network learning approach, that combines secure multi-party computation (MPC), differential privacy (DP) and secret sharing. Since the MPC protocol is implemented via a garbled circuit whose size is subject to the number of parameters (i.e., the size of the gradient) of the neural network, it tends to be less efficient and not scalable while training larger neural networks. Also, in \cite{chase2017private}, using secret sharing requires at least two non-colluding honest data users which might not be practical. 

To address the challenges mentioned above, in this paper, we propose LDP-DL, a privacy-preserving distributed deep learning framework via local differential privacy \cite{wang2019collecting} and knowledge distillation \cite{hinton2015distilling}. Our approach adopts the same ``teacher-student'' paradigm as described in PATE \cite{papernot2016semi}, where each data owner learns a teacher model using its own (local) private dataset, and the data user aims to learn a student model to mimic the output of the ensemble of the teacher models using the unlabelled public data. Knowledge distillation \cite{hinton2015distilling} has been applied on the ensemble of the teacher models to enable faster and more accurate knowledge transferring to the student model, and leverage the advantage of having multiple data owners (teacher models). 
To ensure privacy, our approach employs local differential privacy on the data owners' side, i.e., the query results of each teacher model, which does not require any trusted aggregator (compared to \cite{papernot2016semi}). Since more queries to the teacher models tends to result in more privacy leakage (i.e., cost more privacy budget), we also design an active query sampling approach that could actively select a subset of the unlabelled public dataset for the data user to query from the data owners. In the experimental evaluation, a comprehensive comparison has been made  among our proposed approach (i.e., LDP-DL), DP-SGD \cite{abadi2016deep}, PATE \cite{papernot2016semi} and DP-FL \cite{geyer2017differentially}, using three popular deep learning benchmark datasets (i.e., CIFAR10 \cite{krizhevsky2009learning}, MNIST \cite{lecun1998gradient} and FashionMNIST \cite{xiao2017fashion}). The experimental results show that our LDP-DL framework consistently outperforms the other competitors in terms of privacy budget and model accuracy.

To summarize, our work has the following contributions:
%\begin{itemize}

$\bullet$ We present a novel, effective and efficient privacy-preserving distributed deep learning framework using local differential privacy and knowledge distillation.

$\bullet$ We present an active sampling approach to efficiently reduce the total number of queries from the data user to each data owners, so that to reduce the total cost of privacy budget.

$\bullet$ A comprehensive experimental evaluation among our approach, DP-SGD \cite{abadi2016deep}, PATE \cite{papernot2016semi} and DP-FL \cite{geyer2017differentially} has been conducted on three benchmark dataset (i.e., CIFAR10 \cite{krizhevsky2009learning}, MNIST \cite{lecun1998gradient} and FashionMNIST \cite{xiao2017fashion}). For the sake of reproducibility and convenience of future studies about privacy-preserving distributed deep learning, we have released our prototype implementation of LDP-DL, information regarding the experiment datasets and the code of our comparison experiments. \footnote[1]{\url{https://github.com/nogrady/LDP-DL}}

The rest of this paper is organized as follows: 
Section~\ref{sec:preliminaries} presents the preliminaries including local differential privacy %, LDP mechanisms for multidimensional data 
and knowledge distillation.
Section~\ref{sec:ldp_methodology} presents the problem statement and notations of privacy-preserving distributed deep learning, and describes our proposed framework. 
Section~\ref{sec:ldp_experimentalEvaluation} presents the experimental evaluation. 
Section~\ref{sec:ldp_relatedWork} presents the related literature review. 
Section~\ref{sec:ldp_conclusion} concludes.

\section{Preliminaries} \label{sec:preliminaries}
\subsection{Local Differential Privacy} \label{sec:preliminaries_ldp} 
Differential Privacy (DP) \cite{cynthia2006differential, dwork2014algorithmic} aims to protect the privacy of individuals while releasing aggregated information about the database, which prevents membership inference attacks \cite{shokri2017membership} by adding randomness to the algorithm outcome. Two databases $D$ and $D^{\prime}$ are neighbors if they differ in only one entry. The formal definition is given as follows: 

\begin{definition}
$(\epsilon, \delta)$-Differential Privacy \cite{cynthia2006differential, dwork2014analyze}: A randomized mechanism $A$ is $(\epsilon, \delta)$-differentially private if for every two neighboring databases $D$, $D^{\prime}$ and for any subset $S \subseteq Range(A)$: 
\begin{equation}
Pr[A(D) \in S] \leq e^{\epsilon} \cdot Pr[A(D^{\prime}) \in S] + \delta
\end{equation}
where $Pr[\cdot]$ denotes the the probability of an event, $Range(A)$ denotes the set of all possible outputs of algorithm $A$. Smaller values of $\epsilon$, $\delta$ indicates closer between $Pr[A(D) \in S]$ and $Pr[A(D^{\prime}) \in S]$, thus stronger privacy protection gains. When $\delta=0$ the mechanism $A$ satisfies $\epsilon$-DP, which provides stronger privacy guarantee than $(\epsilon, \delta)$- DP, where $\delta>0$.
\end{definition}

Local Differential Privacy (LDP) \cite{wang2019collecting} is the local setting of DP, which does not require any trusted aggregator. In LDP, individuals (i.e., data owners) send their data to the data aggregator after privatizing data by perturbation. Hence, these techniques provide plausible deniability for individuals (i.e., data owners). Data aggregator collects all perturbed values and makes an estimation of statistics such as the frequency of each value in the population. The formal definition is given as follows: 
\begin{definition}
$\epsilon$-Local Differential Privacy \cite{duchi2013local, wang2017locally}: A randomized mechanism $A$ satisfies $\epsilon$-LDP if for any input $v_{1}$, $v_{2}$ and for any subset $S \subseteq Range(A)$: 
\begin{equation}
Pr[A(v_{1}) \in S] \leq e^{\epsilon} \cdot Pr[A(v_{1}) \in S]
\end{equation}
\end{definition}

Compared with DP, LDP provides more protection to the data owners. Other than sending the private data directly to a trusted aggregator, the data owners could perturb their private data with the mechanism that satisfies $\epsilon$-LDP, and then release the perturbed data. As such, LDP provides a stronger privacy protection, since the aggregator (i.e., data user) only receives the perturbed data and the true values of the private data never leave the hands of the data owners.

%\subsection{LDP Mechanisms for Multidimensional Data} \label{sec:preliminaries_ldp_md}

\subsection{Knowledge Distillation} \label{sec:preliminaries_kd}
Knowledge Distillation (KD) \cite{ba2014deep, hinton2015distilling, polino2018model} was originally designed for deep neural network (DNN) compression and knowledge transfer. KD usually considers a ``teacher-student'' paradigm, where the teacher model is a DNN (or an ensemble of a set of DNNs) that performs well on a given dataset, and the student model is another neural network that may or may not have the same architecture as the teacher model, but aims to mimic the performance of the teacher model(s) using another public dataset. 
Hinton, et al. \cite{hinton2015distilling} proposes an end-to-end knowledge distillation framework with a loss function, namely Distillation Loss, where the output of the teacher model is used as the soft target (i.e., soft label) for the student model, and the overall loss function is presented as below: 
\begin{equation}
\begin{split}
\mathcal{L}(x; \Theta) &= \alpha \cdot \mathcal{H}(y, \sigma (z_{s}; T=1)) \\
&+\beta \cdot \mathcal{H}(\sigma (z_{t}; T=\tau), \sigma (z_{s}; T=\tau))
\end{split}
\label{eq:kd}
\end{equation}
where $y$ is the true label of data $x$, $z_{s}$ is the output of the student model, $z_{t}$ is the output of the teacher model, $\sigma (z_{s}; T=1))$ is the softened label of $z_{s}$ at temperature $T=1$, and $\sigma (\cdot; T=\tau))$ is the softened label at temperature $T=\tau$, and usually $\tau>1$.

\section{Methodology} \label{sec:ldp_methodology}
\subsection{Problem Statement}  \label{sec:ldp_methodology_ps}
\begin{figure}[!h]
  \centering
  \includegraphics[width=0.95\linewidth]{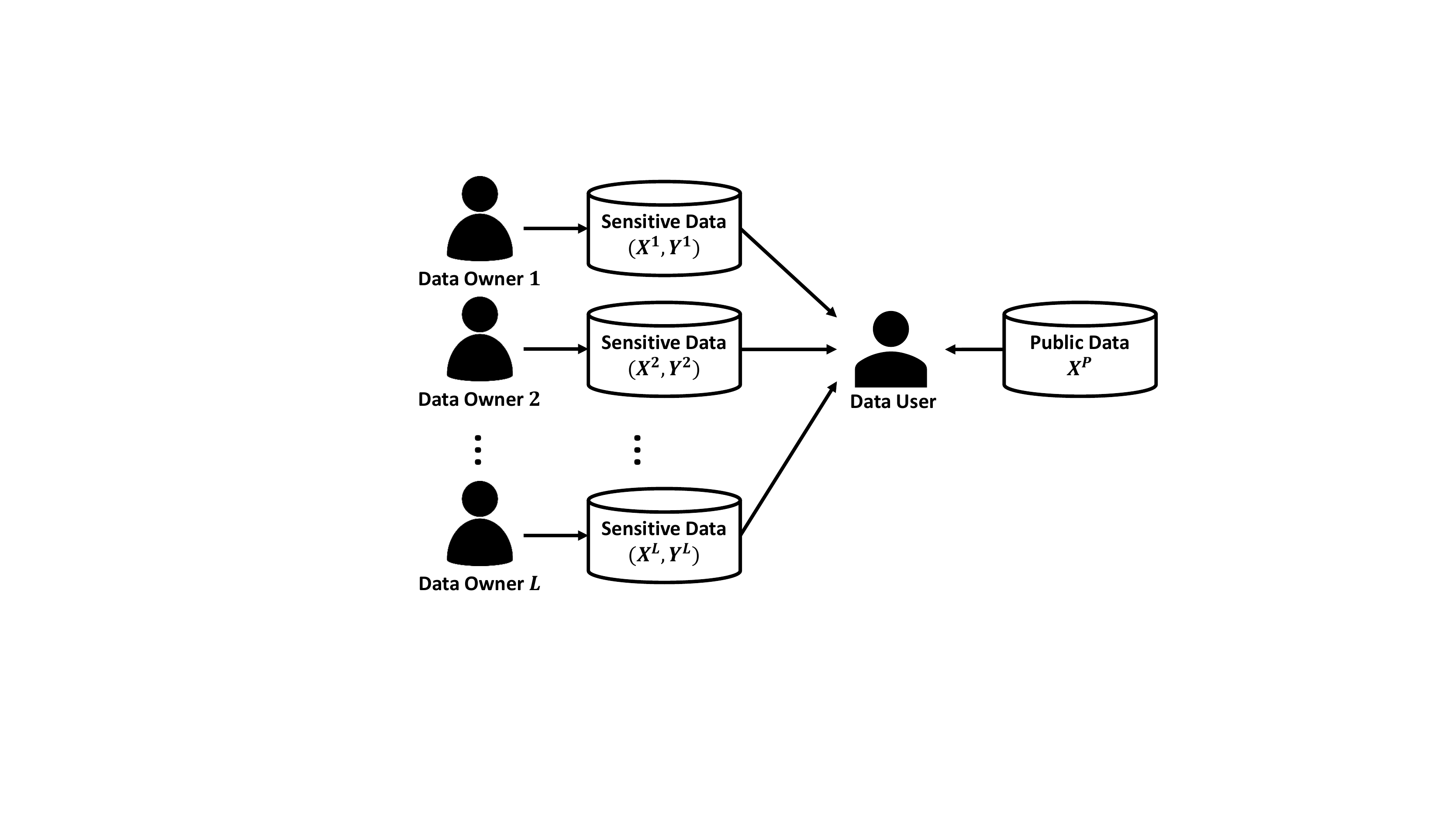}
  \caption{Problem Overview.}
  \label{fig:methodology_ps}
\end{figure}

In this work, we aim to develop a privacy-preserving distributed deep learning framework. As shown in Fig.~\ref{fig:methodology_ps}, we consider the following problem: Given $L$ data owners, each data owner $l$ holds a set of private samples $(X^{l}, Y^{l})$, where $X^{l}=\{x_{1}^{l}, x_{2}^{l}, \dots, x_{n^{l}}^{l}\}$, $Y^{l}=\{y_{1}^{l}, y_{2}^{l}, \dots, y_{n^{l}}^{l}\}$, $x_{i}^{l} \in \mathbb{R}^{d}$, and $y_{i}^{l} \in \{1, 2, \dots, k\}$ is the label associated with sample $x_{i}^{l}$, $i = 1, 2, \dots, n^{l}$; the untrustworthy data user would like to learn a DNN model with the help of all the data owners, and a public dataset $X^{P}$ that comes from the same distribution (i.e., the same problem) as the data owners' private datasets, but does not have the label information. In our problem setting, each data owner has two privacy requirements: (i) the value of the individual private data should not be shared to the data user, and (ii) any inference of the individual private data should be prevented from using the intermediate communication messages and the data user's DNN model.

\subsection{Threat Model}  \label{sec:ldp_methodology_tm} 
In our problem, we assume (i) the data user is untrustworthy, and (ii) the data owners are honest-but-curious, where each data owner follows the protocol honestly, but try to use the protocol transcripts to extract new information. We assume the value of the individual private data is what the adversaries would like to acquire during the whole protocol. Hence, the adversaries could be the data user, the participating data owners or an outside attacker that has the access to the intermediate communication messages or the data user's DNN model. We also assume that the adversaries may have arbitrary background knowledge and might collude with each other. 
Our work aims to protect the privacy of each data owner's individual private data while providing the reasonable utility to the data user's DNN model. Since we assume that the data user is untrustworthy, it is of the data user’s own interest to correctly execute the algorithm or not. However, while using our proposed framework, if the untrustworthy data user behave dishonestly, it will not compromise data owner’s privacy, but will only hurt the utility of the data user's DNN model. Furthermore, since the data owners are assumed to be honest-but-curious, the poisoning \cite{biggio2012poisoning}, backdoor \cite{gu2017badnets, chen2017targeted} or trojans \cite{liu2017neural} attacks (e.g., the data owner actively and maliciously modify their inputs to influence the performance of the data user's DNN model) are beyond the scope of this work.

\begin{figure*}[!h]
  \centering
  \includegraphics[width=0.95\linewidth]{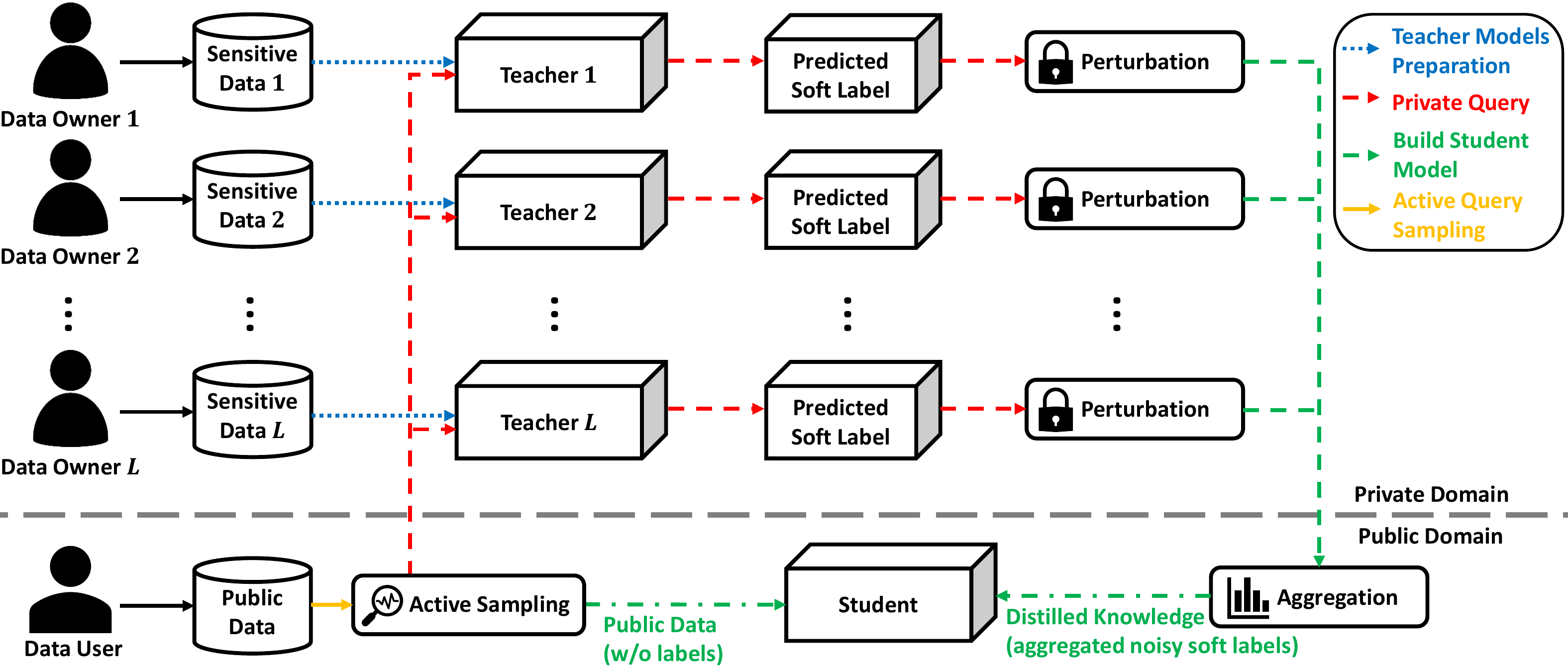}
  \caption{The Overview of LDP-DL Framework.}
  \label{fig:Framework}
\end{figure*}

\begin{algorithm}[!h]
\caption{Locally Differentially Private Distributed Deep Learning (LDP-DL)}\label{alg:LDP_DL}
\LinesNumbered
\KwIn{$\{(X^{1}, Y^{1}), (X^{2}, Y^{2}), \dots, (X^{L}, Y^{L})\}$; $X^{P}$; active query sampling size (per iteration) $S$; the number of queries available for each selected public sample $N_{Q}$; distillation learning batch size $S_{db}$.}
\KwOut{Student Model $M_{S}$.}

\For {$l \in \{1, 2, \dots, L\}$}{
    Data owner $l$ trains a teacher model $M_{T_{l}}$ using his/her private dataset $(X^{l}, Y^{l})$\;
}

Initialize the student model $M_{S}$ by the data user\;

\While{$|X^{P}| \geq S$ \text{and} the accuracy of $M_{S}$ is not acceptable}{
    $X^{Q} \longleftarrow$ $ActiveQuerySampling(M_{S}, X^{P}, S)$\; 
    $X^{P} \longleftarrow X^{P}-X^{Q}$\;
    
    $(X^{Q^{\prime}}, Y^{Q^{\prime}})=\O$\;
    \For {$x \in X^{Q}$}{
        $z_{t} \longleftarrow (0, 0, \dots, 0)^{1 \times k}$\;
        
        Randomly select $N_{Q}$ data owners from $\{1, 2, \dots, L\}$ to form a subset $O_{Q}$\;
        
        \For {$l \in O_{Q}$}{
            $z_{t}^{l} \longleftarrow PrivacySanitize(M_{T_{l}}(x))$\;
            $z_{t} \longleftarrow z_{t} + z_{t}^{l}$\;
        }
        $z_{t} \longleftarrow z_{t} / N_{Q}$\;
        
        $(X^{Q^{\prime}}, Y^{Q^{\prime}}) \longleftarrow (X^{Q^{\prime}}, Y^{Q^{\prime}}) \cup \{(x, z_{t})\}$\;
    }
    
    \While{$|(X^{Q^{\prime}}, Y^{Q^{\prime}})| \geq S_{db}$}{
        Sample a batch $(X^{Q^{\prime}}_{i}, Y^{Q^{\prime}}_{i})$ of size $S_{db}$ from $(X^{Q^{\prime}}, Y^{Q^{\prime}})$\; 
        Compute distill loss $\mathcal{L}_{distill}^{i}$ over $(X^{Q^{\prime}}_{i}, Y^{Q^{\prime}}_{i})$ by Eq.\;
        Backpropagate by $\mathcal{L}_{distill}^{i}$ to update $M_{S}$\;
    }
}

\Return $M_{S}$.
\end{algorithm}

\subsection{Privacy-preserving Distributed Deep Learning}  \label{sec:ldp_methodology_ldp_dl}
Our proposed privacy-preserving distributed algorithm framework, as shown in Fig.~\ref{fig:Framework}, consists of four stages that work synergistically between the data owners and the data user. Alg.\ref{alg:LDP_DL} shows the pseudo-code of our algorithm. 
Firstly, each data owner trains a teacher model using his/her own private dataset (i.e., lines 1-2), and the data user initialize the student models with random or pretrained (i.e., ImageNet \cite{deng2009imagenet}) parameters (i.e., line 3). The student model and all the teacher models do not have to use the same DNN architecture. 
Secondly, in each iteration, the data user efficiently selects a subset of the available public dataset (i.e., lines 5-6), that could better improve the performance of the current student model in the upcoming training, using our well-designed active query sampling approach. The active query sampling component aims to reduce the total number of queries to each teacher model, thus saving the privacy budgets. 
Thirdly, the data user uses the selected subset of the public data (no labels) to query each of the teacher model from its corresponding selected data owner to obtain the ``knoweledge'' (data's soft label), and all the query results are sanitised by our local differential privacy techniques before being sent back to the data user (i.e., lines 7-15). Since the data owner might select a huge amount of query samples, it is not realistic to use all the selected samples to query all the data owners, which cost much on the privacy budge and the communication, but might not help a lot for the utility (per our experimental results). Therefore, we predefined a parameter $N_{Q}$ to control/specify the upper bound of the available data owners for each selected public sample to query (lines 10-11). 
Last but not least, the data user aggregates the received sanitised query results (i.e., the distilled knowledge) of each data, and leverage the knowledge distillation techniques (using the subset of the public data, and the distilled knowledge) to update/train the student model. The details of the most important three components in our framework (i.e, private query from teacher models, build student model via knowledge transfer and active query sampling) are described in the subsequent sections.

%\subsubsection{Teacher Models Preparation by Data Owners}  \label{sec:ldp_methodology_teacher}

\begin{algorithm}[!h]
\caption{Piecewise Mechanism for One-Dimensional
Numerical Data (PM-ONE) \cite{wang2019collecting}}\label{alg:PM_ONE}
\LinesNumbered
\KwIn{tuple $z_{i} \in [-1, 1]$; privacy budget $\epsilon$.}
\KwOut{perturbed tuple $z^{\prime}_{i} \in [-\Delta, \Delta]$.}

$\Delta \longleftarrow \frac{e^{\epsilon/2}+1}{e^{\epsilon/2}-1}$\;
$L(z_{i}) \longleftarrow \frac{\Delta+1}{2} \cdot z_{i} - \frac{\Delta-1}{2}$\;
$R(z_{i}) \longleftarrow L(z_{i}) +\Delta - 1$\;
Sample value $v$ uniformly at random from $[0, 1]$\;

\If {$v<\frac{e^{\epsilon/2}}{e^{\epsilon/2}+1}$} {
    Sample $z_{i}^{\prime}$ uniformly at random from $[L(z_{i}), R(z_{i})]$\;
}
\Else{
     Sample $z_{i}^{\prime}$ uniformly at random from $[-\Delta, L(z_{i})] \cup [R(z_{i}), \Delta]$\;
}
\Return $z_{i}^{\prime}$.
\end{algorithm}

\begin{algorithm}[!h]
\caption{Piecewise Mechanism for Multidimensional
Numerical Data (PM) \cite{wang2019collecting}}\label{alg:PM}
\LinesNumbered
\KwIn{tuple $z \in [-1, 1]^{k}$; privacy budget $\epsilon$.}
\KwOut{perturbed tuple $z^{\prime} \in [-k \cdot \Delta, k \cdot \Delta]^{k}$.}

$z^{\prime} \longleftarrow <0, 0, \dots, 0>$\;

$m \longleftarrow max\{1, min\{k, \lfloor \frac{\epsilon}{2.5} \rfloor\}\}$\;

Sample $m$ values uniformly without replacement from $\{1, 2, \dots, k\}$\;

\For {\text{each sampled attribute} $j$}{
    $z^{\prime}_{j}=\frac{k}{m} \cdot PM$-$ONE(z_{j}, \frac{\epsilon}{m})$\;
}

\Return $z^{\prime}$.
\end{algorithm}

\subsection{Private Query from Teacher Models} \label{sec:ldp_methodology_query}
In our proposed algorithm, upon receiving the query data from the data user, each data owner evaluates it using his/her own teacher model and gets the data's soft label. Each data owner perturbs the query data's soft label (using LDP techniques) and then sends the perturbed value to the data user to transfer the distilled knowledge. The data user then aggregates the perturbed query results of each data to obtain the aggregated noisy soft label (i.e., averaged over the query results sent by all the selected data owners) of each data. As such, we could formulate this as a locally differentially private mean estimation problem, where we would like to protect the data owners' private data from the inference attacks given the perturbed query results, and ensure the aggregated noisy soft label as close as the real value. 
As described in Section~\ref{sec:preliminaries_ldp}, while applying LDP, the adversaries could not distinguish the true value from a perturbed value with a high confidence (adjusted by the parameter $\epsilon$). To protect the privacy
of the data owners' private data, the randomization method which satisfies $\epsilon$-LDP is adopted. On the other hand, the performance of the aggregation of the perturbed data could be maintained with an error bound \cite{wang2017locally, duchi2018minimax}, which provides us a way to control the utility of the distilled knowledge. Furthermore, since all the soft labels are multidimensional numerical values, different from the hard labels which are categorical values, we can not directly adopt the encoding-based LDP techniques \cite{wang2017locally}.  

To achieve our goal, we adopt the Piecewise Mechanism (PM) \cite{wang2019collecting} that is designed to perturbed the multidimensional numerical values, and has an asymptotic optimal error bound for the mean estimation problem. 
Alg.~\ref{alg:PM_ONE} shows the PM for one-dimensional numerical data (i.e., PM-ONE). To simplify our explanation, in this section, the value to be perturbed (i.e., the soft labels of $k$ classes) is denoted as $z \in \mathbb{R}^{k}$, $z=[z_{1}, z_{2}, \dots, z_{k}]$. PM-ONE (Alg.~\ref{alg:PM_ONE}) takes a one-dimentional numerical data $z_{i} \in [-1, 1]$ as the input, and returns its perturbed value $z^{\prime}_{i} \in [-\Delta, \Delta]$, where $\Delta \leftarrow \frac{e^{\epsilon/2}+1}{e^{\epsilon/2}-1}$ is small and thus $z^{\prime}_{i}$ has relatively high probability (i.e., $\frac{e^{\epsilon/2}}{e^{\epsilon/2}+1}$) to be close to $z_{i}$. As shown in \cite{wang2019collecting}, while applying PM-ONE $|\frac{1}{n}\sum_{i=1}^{n} z_{i}^{\prime}-\frac{1}{n}\sum_{i=1}^{n} z_{i}|=O(\frac{\sqrt{log(1/\beta)}}{\epsilon\sqrt{n}})$ with at least $1-\beta$ probability for the task of mean estimation, which is an asymptotically optimal error bound.

Alg.~\ref{alg:PM} shows the PM for multidimensional numerical data (i.e., PM), where for each data of $k$ dimensions, it randomly selects $m$ (i.e., $m<k$) attributes to perturb. Alg.~\ref{alg:PM} is designed to reduce the amount of the noise in the task of mean estimation for multidimensional numerical data. While using Alg.~\ref{alg:PM_ONE} to perturb $k$ attributes, each attribute evenly shares a privacy budget of $\frac{\epsilon}{k}$, and the total amount of noise in the mean estimation is $O(\frac{k\sqrt{log(k)}}{\epsilon \sqrt{n}})$, which is super-liner to $k$. However, it has been shown \cite{wang2019collecting} that while using Alg.~\ref{alg:PM}, $E[ max_{j \in [1, k]} |\frac{1}{n}\sum_{i=1}^{n} z_{i,j}^{\prime}-\frac{1}{n}\sum_{i=1}^{n} z_{i,j}|]=O(\frac{\sqrt{klog(k/\beta)}}{\epsilon\sqrt{n}})$ with at least $1-\beta$ probability for the task of mean estimation, which is still an asymptotically optimal error bound. 

\subsubsection{Privacy Budget Analysis of LDP-DL} \label{sec:Privacy_Budget_Analysis}
As described in Section~\ref{sec:ldp_methodology_ps}, in our proposed framework, there are $|X^{P}|$ public data and $L$ data owners in total. If each public data could query the teacher model for at most $N_{Q}$ times (Section~\ref{sec:ldp_methodology_ldp_dl}), each teacher model will be queried for at most $r=\frac{|X^{P}| \cdot N_{Q}}{L}$ times by average. Suppose for each private query, the perturbed query result satisfies $\epsilon_{i}$-LDP. According to the composition property of LDP \cite{mcsherry2007mechanism}, to meet the requirement of $\epsilon$-LDP for each data owner's private data, we need to satisfy $\sum_{i}^{r} \epsilon_{i} \leq \epsilon$. Since each data owner would participate in the private query for at most $r$ times by average, it requires $\epsilon_{i} \leq \frac{\epsilon}{r}$. Then, the noise of each query result becomes $O(\frac{r\sqrt{klog(k)}}{\epsilon})$, which is linear to $r$. Since each public data would be queried for at most $N_{Q}$ times, the noise of the mean estimation of each public data's soft label (i.e., distilled knowledge) would be $O(\frac{r\sqrt{klog(k)}}{\epsilon\sqrt{N_{Q}}})=O(\frac{|X^{P}| \cdot \sqrt{N_{Q}} \cdot \sqrt{klog(k)}}{\epsilon \cdot L} )$. Since $\epsilon$ is the privacy budget that should be controlled by the data owner's preference, and $N_{Q}$ has the direct influence on the precision of each query result's mean estimation that should be decided on the data user's empirical study, to reduce the overall noise, it is better to increase the number of participant data owners (i.e., $L$) or decrease the size of the set of public data (i.e., $|X^{P}|$) utilized for private query (as described in Section~\ref{sec:ldp_methodology_sampling}).

\subsection{Build Student Model via Knowledge Transfer}  \label{sec:ldp_methodology_student}
While the data user receiving all the query results from the data owners, our proposed framework uses the knowledge distillation technique to transfer the knowledge learned from the queried teacher models to the student model. Our usage of knowledge distillation is slightly different from the it convectional usage (as described in Section~\ref{sec:preliminaries_kd}), where (i) we only focus on the knowledge transfer perspective of KD, but not the model compression, thus the student model and all the teacher models could use different and arbitrary DNN architectures; and (ii) in our case, the public dataset does not have the true label information, thus cannot directly use equation~(\ref{eq:kd}). 
Hence, in our framework, the student model is trained to minimize the gap between its own predicted soft label and the aggregated soft label from the teacher models, i.e., the knowledge distillation loss:
\begin{equation}
\begin{split}
\mathcal{L}(x; \Theta) &= \alpha \cdot \mathcal{H}(\sigma (z_{t}; T=1), \sigma (z_{s}; T=1)) \\
&+\beta \cdot \mathcal{H}(\sigma (z_{t}; T=\tau), \sigma (z_{s}; T=\tau))
\end{split}
\label{eq:kd_ours}
\end{equation}
where $z_{s}$ is the soft label predicted by the student model, $z_{t}$ is the aggregated soft label from the teacher models, $T$ is the temperature parameter, and $\sigma (z; T) = softmax (z/T)$. The temperature parameter is usually set to 1. While $T>1$, the probabilities of the classes whose normal values are near zero would be increased. To better distill the knowledge to the student, two temperature values are adopted in our KD loss (i.e., $T=1$ and $T=\tau>1$).

\begin{figure}[!h]
  \centering
  \includegraphics[width=0.95\linewidth]{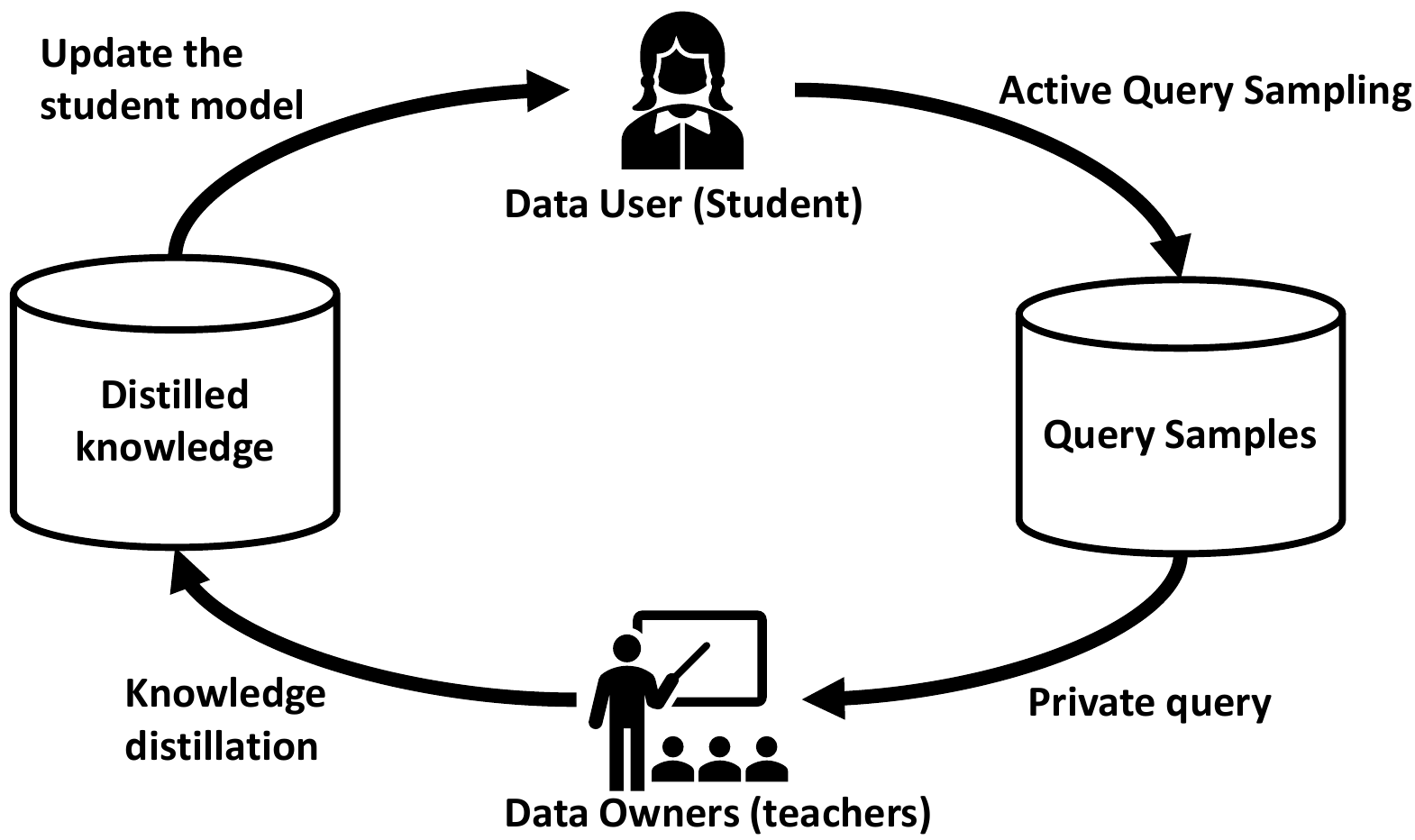}
  \caption{Active Query Sampling.}
  \label{fig:AL}
\end{figure}

\subsection{Active Query Sampling}  \label{sec:ldp_methodology_sampling}
As analyzed in Section~\ref{sec:Privacy_Budget_Analysis}, one direction to reduce the overall noise of the soft label estimation (thus, enhance the overall performance) is to decrease the size of the set of public data (i.e., $|X^{P}|$) utilized for private query. In this section, we present an active query sampling approach that could actively and adaptively choose samples from the public dataset batch-by-batch to query the teacher models. As shown in Fig.~\ref{fig:AL}, we adopt a ``least confidence'' strategy \cite{settles2009active}, where in iteration, we attempt to select a set of query samples from the public dataset that the student model shows the ``least confidence'' at. To be specific, our active query sampling follows the procedure described below:
\begin{enumerate}
    \item Select an initial subset of $S$ unlabeled public data $X^{Q}$  uniformly at random from $X^{P}$. Update $X^{P} \leftarrow X^{P}-X^{Q}$. 
    \item Use $X^{Q}$ to query the teacher models, and use the distilled knowledge to train the student initial student model $M_{S}$. 
    \item For each available public data $x_{i} \in X_{P}$, evaluate it on the student model $M_{S}$. Let $P_{ij}$ denote the probability of $x_{i}$ belonging to class $j \in \{1, 2, \dots, k\}$ predicted by $M_{S}$. Let $P_{i}=\{P_{i1}, P_{i2}, \dots, P_{ik}\}$, and suppose $\sum_{l=1}^{k} P_{il}=1$. Let $P_{i}^{*}$ be the largest value (posterior probability) in $P_{i}$. Then, repeat the procedure below for $S$ times to select $S$ query samples: 
\begin{equation}
\begin{split}
x_{i} &\leftarrow X^{P}\\
P_{i} &\leftarrow M_{S}(x_{i})\\
X^{Q} &\leftarrow X^{Q} \cup \underset{x_{i}}{argmin}\frac{1}{m-1} \sum_{l=1}^{k} (P_{i}^{*}-P_{il}) \\
X^{P} &\leftarrow X^{P}-X^{Q}
\end{split}
\end{equation}
Then, use $X^{Q}$ to query the teacher models, and use the distilled knowledge to train the student initial student model $M_{S}$.
\item Repeat 3), until the student model meet the performance requirement or no more public data available (i.e., $X_{P}=\O$). 
\end{enumerate}

\section{Experimental Evaluation} \label{sec:ldp_experimentalEvaluation}
In this section, we evaluate the effectiveness %and efficiency 
of our proposal method, LDP-DL, on three popular image benchmark datasets (i.e., CIFAR-10~\cite{krizhevsky2009learning}, MNIST~\cite{lecun1998gradient} and Fashion-MNIST~\cite{xiao2017fashion}) with three basic LDP mechanisms (i.e., Piecewise mechanism \cite{wang2019collecting}, Duchi's mechanism \cite{duchi2013local} and Laplace mechanism \cite{yang2012differential}). We also evaluate the performance of our proposed Active Query Sampling (AQS) of our approach. Then, we compare LDP-DL with three state-of-the-art approaches, i.e., DP-SGD~\cite{abadi2016deep}, PATE~\cite{papernot2016semi} and DP-FL~\cite{geyer2017differentially}.

% We first verified the effectiveness of LDP-DL and investigated the effect of different parameter settings. Then, we compared LDP-DL with baseline approaches, including non-active query LDP-DL and 3 state-of-art approaches.

\subsection{Experiment Environment}
All the experiments were conducted on a PC with an Intel Core i9-7980XE processor, 128GB RAM, a Nvidia GeForce GTX 1080Ti graphic card, running 64-bit Ubuntu 18.04 LTS operating system. All the experiments are implemented using Python 3.7.

\subsection{Experiment Datasets}
Three popular benchmark image datasets are utilized to conduct our experimental evaluation:

%Three datasets are used to conduct the experiments:
\begin{enumerate}
    \item CIFAR-10~\cite{krizhevsky2009learning}: CIFAR-10 (Canadian Institute For Advanced Research) is a widely used benchmark dataset to evaluate deep learning algorithms. This dataset is a subset of the 80 million tiny images dataset. It contains 60,000 32 x 32 color photographs of objects in 10 different classes, such as frogs, birds, cats, ships, etc. For each class, there are 6,000 images in total, where the testing set includes exactly 1,000 images that randomly selected from each class, and the training set contains the remaining 5,000 images in a random order.
    
    \item MNIST~\cite{lecun1998gradient}: MNIST (Modified National Institute of Standards and Technology database) dataset is a collection of handwritten digits that is commonly used in the field of image processing and machine learning. This dataset is created by "re-mixing" samples from the NIST dataset. It contains 70,000 28 x 28 grayscale images in 10 different classes, i.e., 10 digits, from 0 to 9. The handwritten digits have been size-normalized and centered in each images. The 70,000 samples have been separated to 60,000 training samples and 10,000 testing samples.
    
%    for training and testing purpose, respectively.
                
    \item Fashion-MNIST~\cite{xiao2017fashion}: This dataset is a collection of Zalando’s article images, which is created as a drop-in (more challenging) replacement for MNIST to better represent modern computer vision tasks. It contains 70,000 28 x 28 gray-scale images in 10 different classes. Each class is a kind of cloth, such as T-shirt, dress, trouser, sneaker, etc. There are 60,000 training samples and 10,000 testing samples.
\end{enumerate}

\subsection{Experimental Setup}
In our experiments, we assume the data owner's teacher models are using ResNet50, and the data user's student model is using ResNet18. For each experiment dataset, we assume each data owner has 4,000 private samples to train his/her teacher model (i.e., ResNet50). The data user would query 200 public samples in each iteration of the Active Query Sampling (AQS) process, and queries 5 iterations in total to train his/her student model (i.e., ResNet18). 
Each teacher and student model has been trained for 20 epochs with a batch size of 32.

As discussed in Section~\ref{sec:Privacy_Budget_Analysis}, there are three major parameters that we would like to tune and evaluate in our approach, such as the privacy budget ($\epsilon$), the number of queries of each public sample ($N_{Q}$) and the total number of participated data owners ($L$). We use various combinations of $\epsilon$, $N_{Q}$ and $L$ to evaluate our approach, where $\epsilon$ $\in$ $\{1$, $2$, $3$, $4$, $5$, $6$, $7$, $8$, $9$, $10\}$, $N_{Q}$ $\in$ $\{10$, $20$, $30$, $40$, $50$, $60$, $70$, $80$, $90$, $100\}$, and  $L$ $\in$ $\{1,000$, $2,000$, $3,000$, $4,000$, $5,000$, $6,000$, $7,000$, $8,000$, $9,000$, $10,000\}$. 

Then, we evaluate the performance of our approach while with and without the Active Query Sampling (AQS) process. Also, we evaluate the performance of our approach's private query using  three basic LDP mechanisms, including Piecewise mechanism \cite{wang2019collecting}, Duchi's mechanism \cite{duchi2013local} and Laplace mechanism \cite{yang2012differential}. All the experiments have been repeated for 10 times and we take the average as the reported results.

\begin{figure*}[h]
\centering
	\begin{subfigure}{.3\textwidth}
		\includegraphics[width=\textwidth]{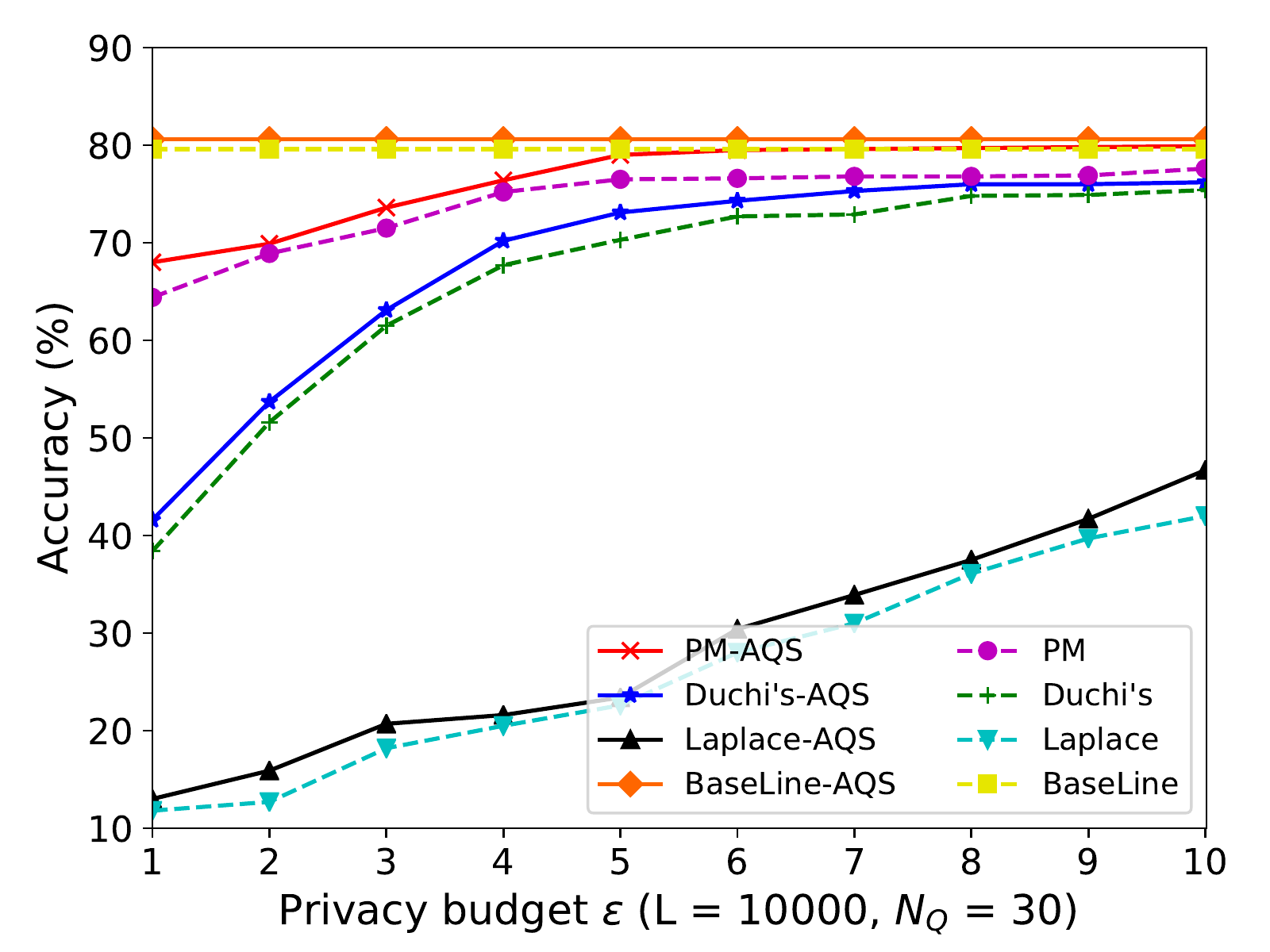}
		\caption{}
        \label{CIFAR10_TotalBudget_smooth}
	\end{subfigure}
%%%%%%%%%%%%%%
	\begin{subfigure}{.3\textwidth}
		\includegraphics[width=\textwidth]{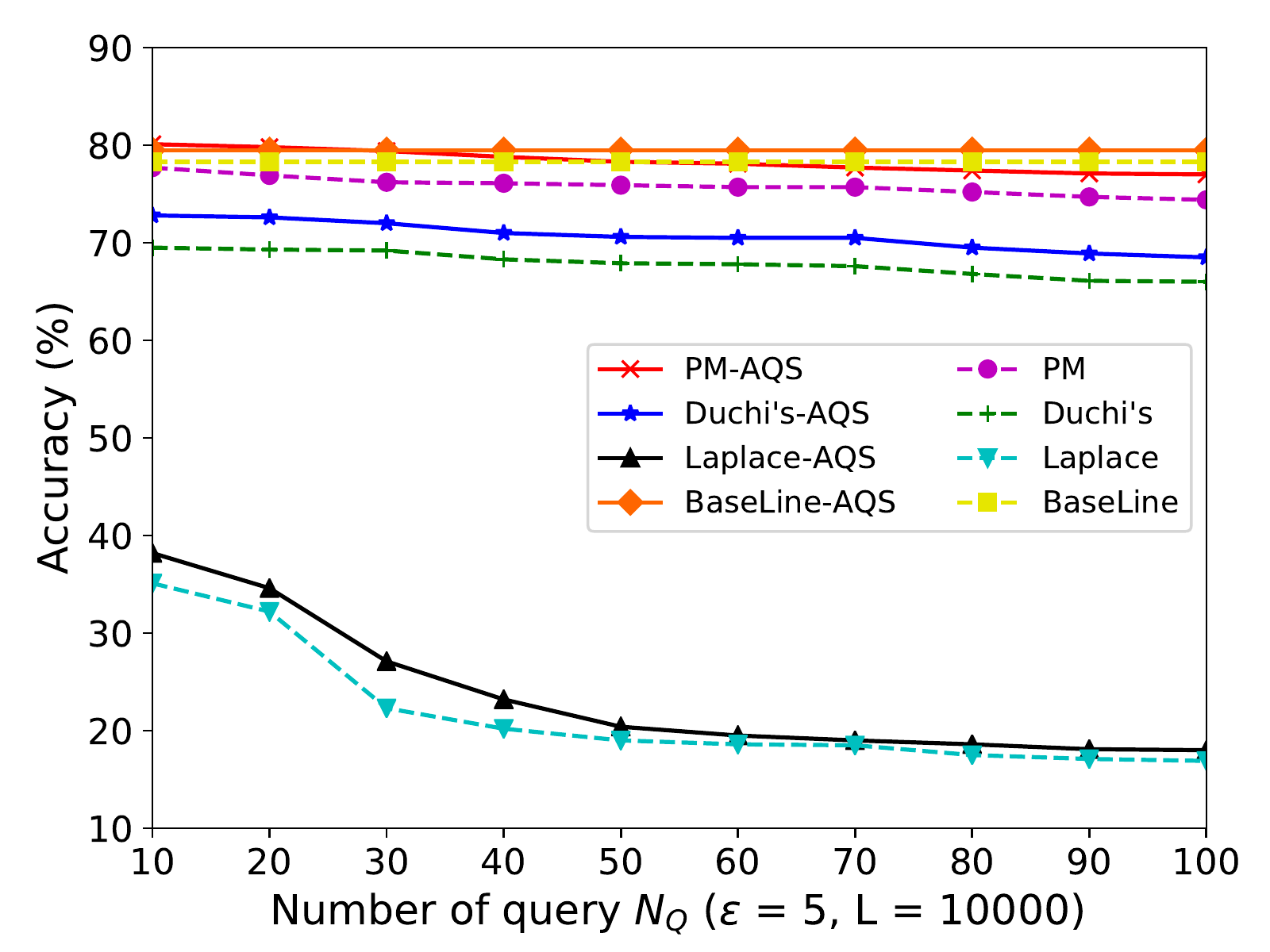}
		\caption{}
        \label{CIFAR10_NumberofQuery_smooth}
	\end{subfigure}
%%%%%%%%%%%%%%
	\begin{subfigure}{.3\textwidth}
		\includegraphics[width=\textwidth]{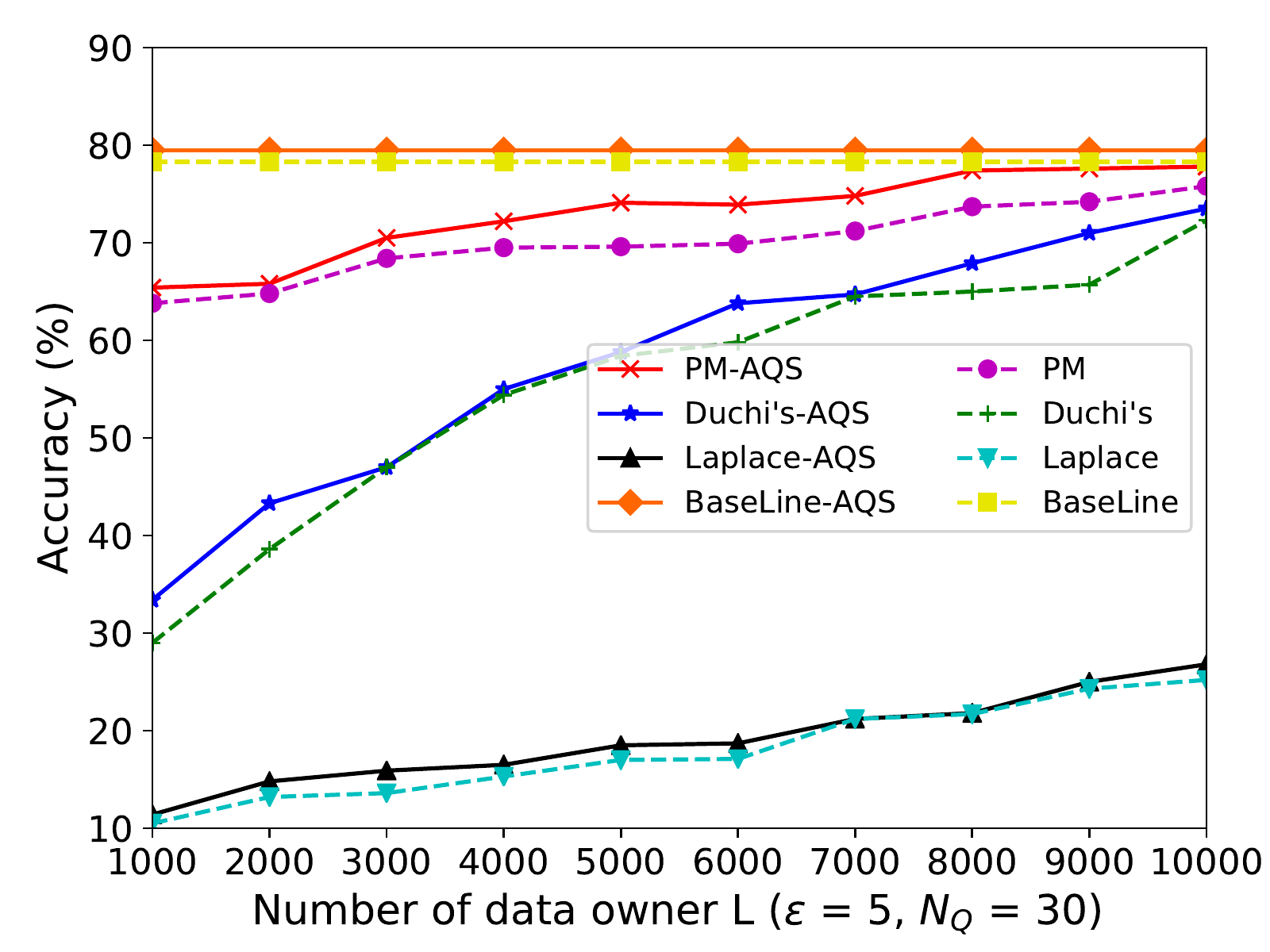}
		\caption{}
        \label{CIFAR10_DataOwner_smooth}
	\end{subfigure}
%%%%%%%%%%%%%%
	\caption{LDP-DL experimental results on CIFAR10 dataset.}
    \label{CIFAR10_dataset}
\end{figure*}

\begin{figure*}[h]
\centering
	\begin{subfigure}{.3\textwidth}
		\includegraphics[width=\textwidth]{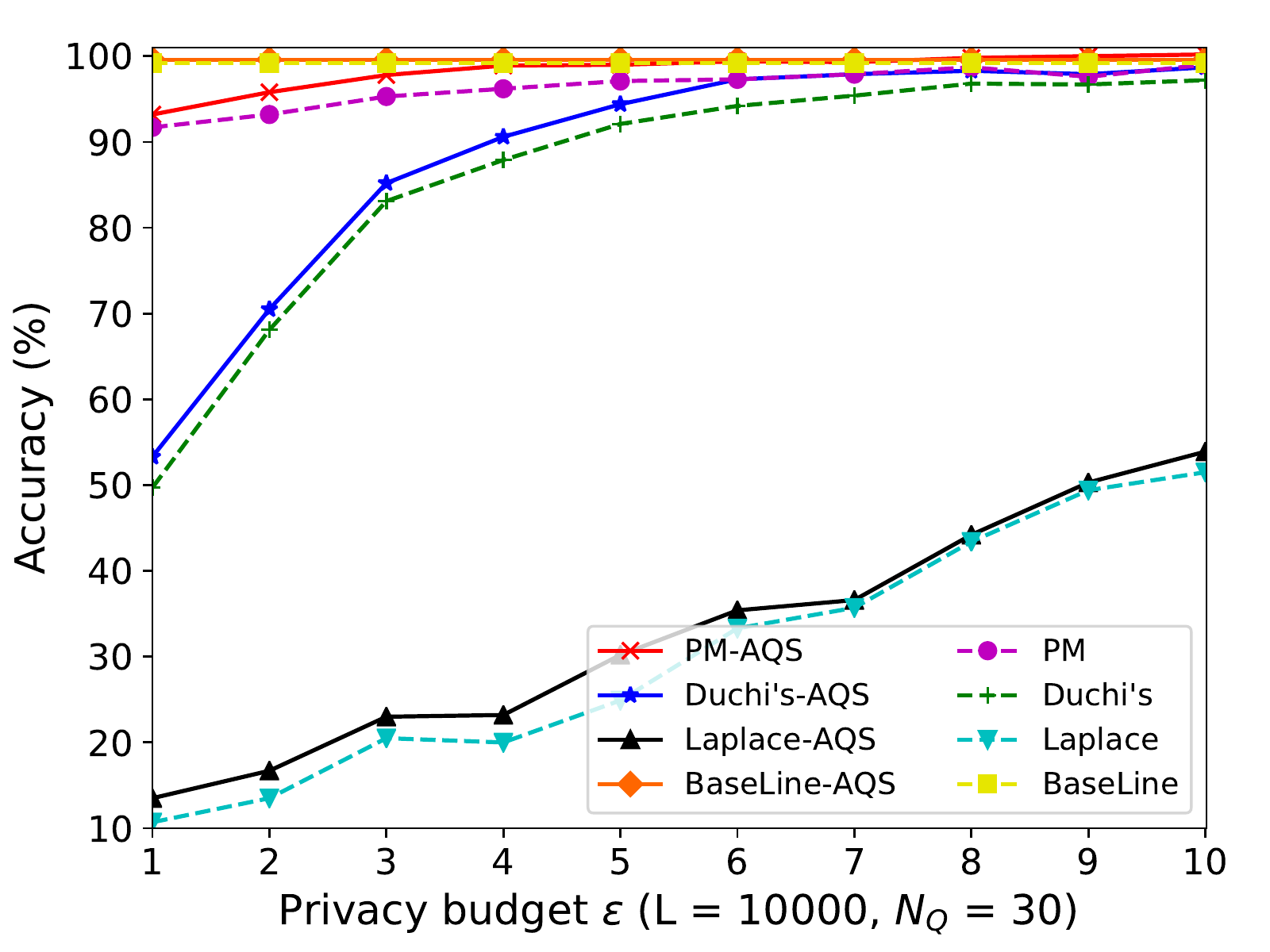}
		\caption{}
        \label{MNIST_TotalBudget_smooth}
	\end{subfigure}
%%%%%%%%%%%%%%
	\begin{subfigure}{.3\textwidth}
		\includegraphics[width=\textwidth]{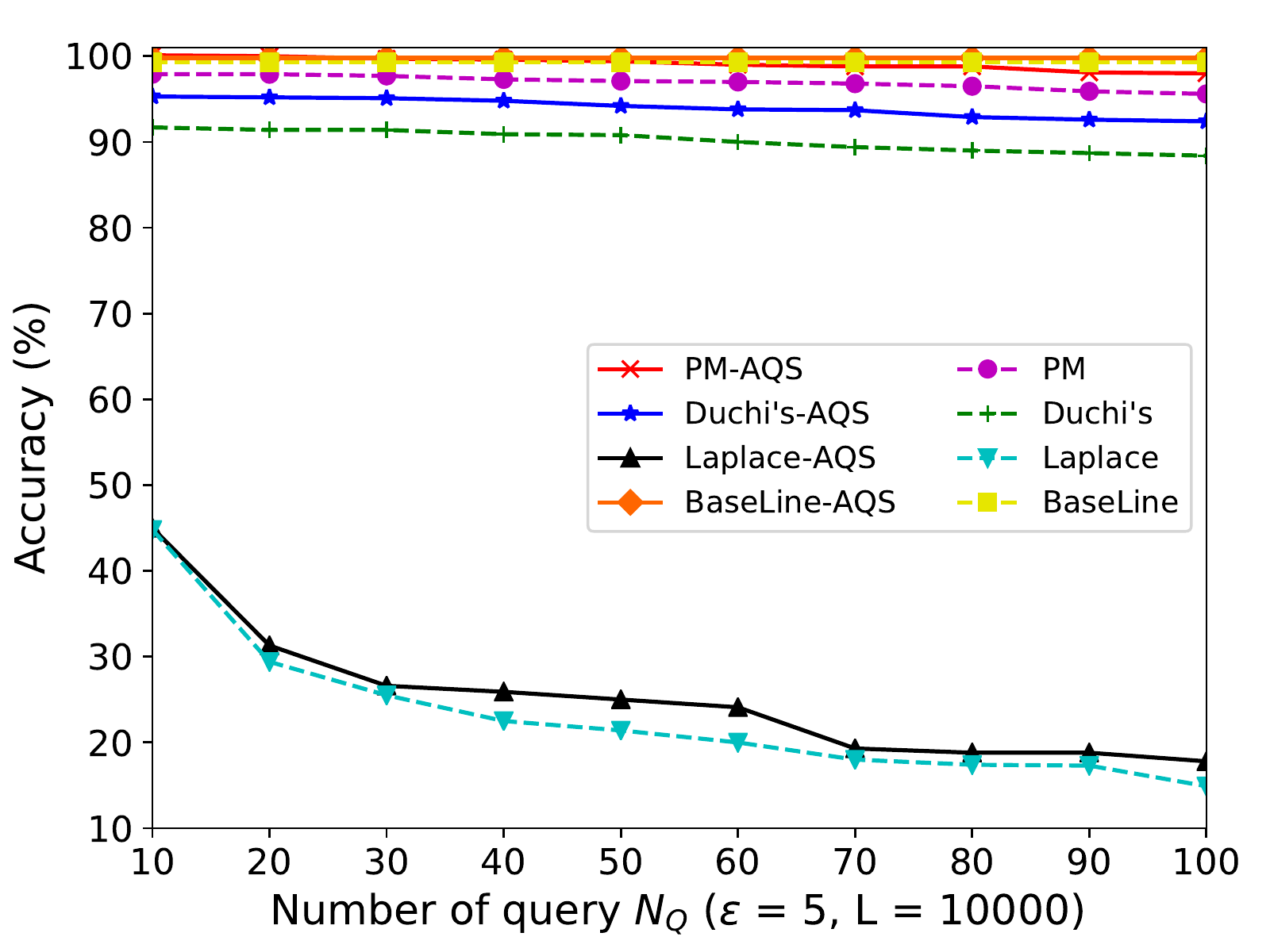}
		\caption{}
        \label{MNIST_NumberofQuery_smooth}
	\end{subfigure}
%%%%%%%%%%%%%%
	\begin{subfigure}{.3\textwidth}
		\includegraphics[width=\textwidth]{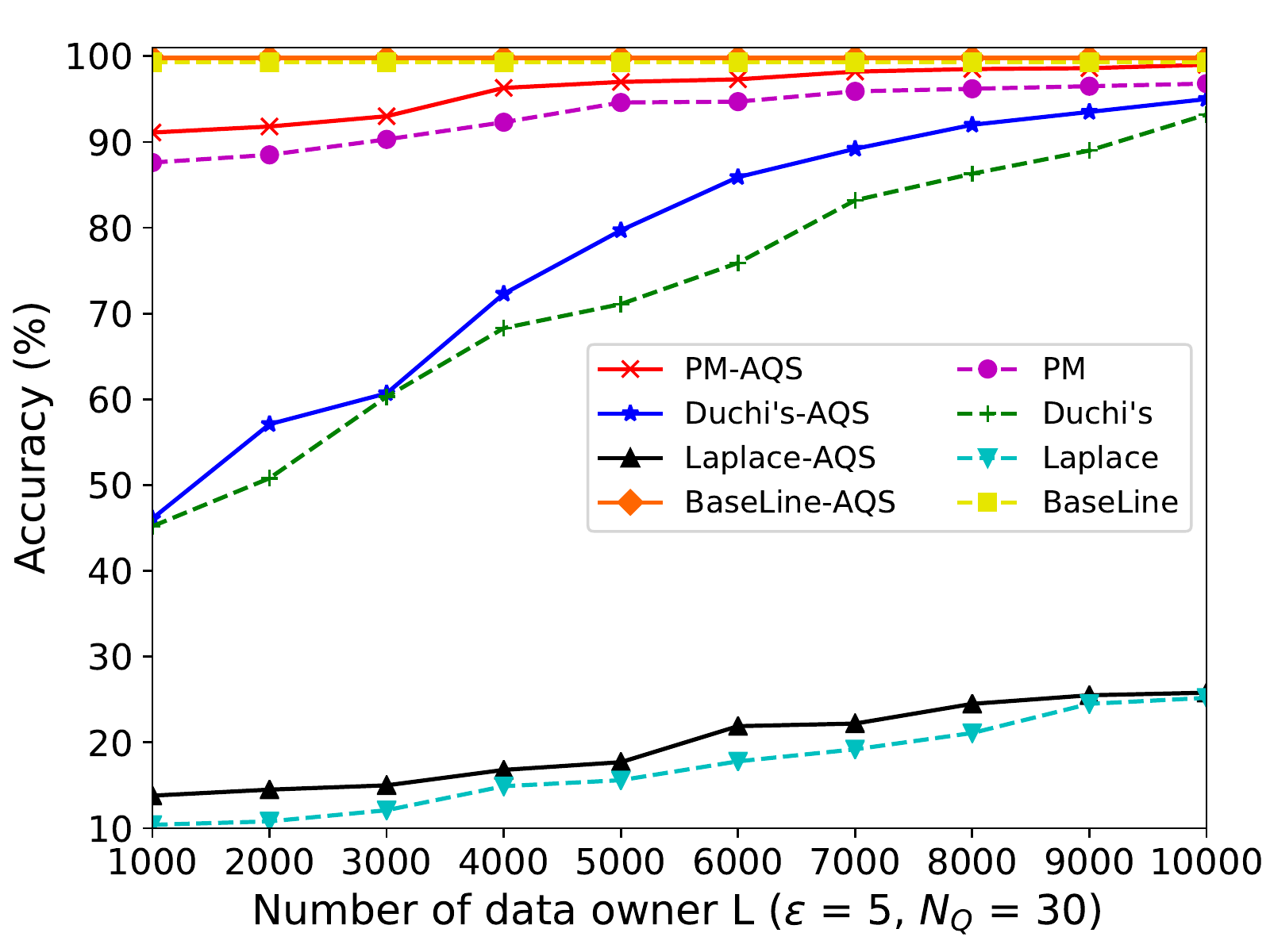}
		\caption{}
        \label{MNIST_DataOwner_smooth}
	\end{subfigure}
%%%%%%%%%%%%%%
	\caption{LDP-DL experimental results on MNIST dataset.}
    \label{MNIST_dataset}
\end{figure*}

\begin{figure*}[h]
\centering
	\begin{subfigure}{.3\textwidth}
		\includegraphics[width=\textwidth]{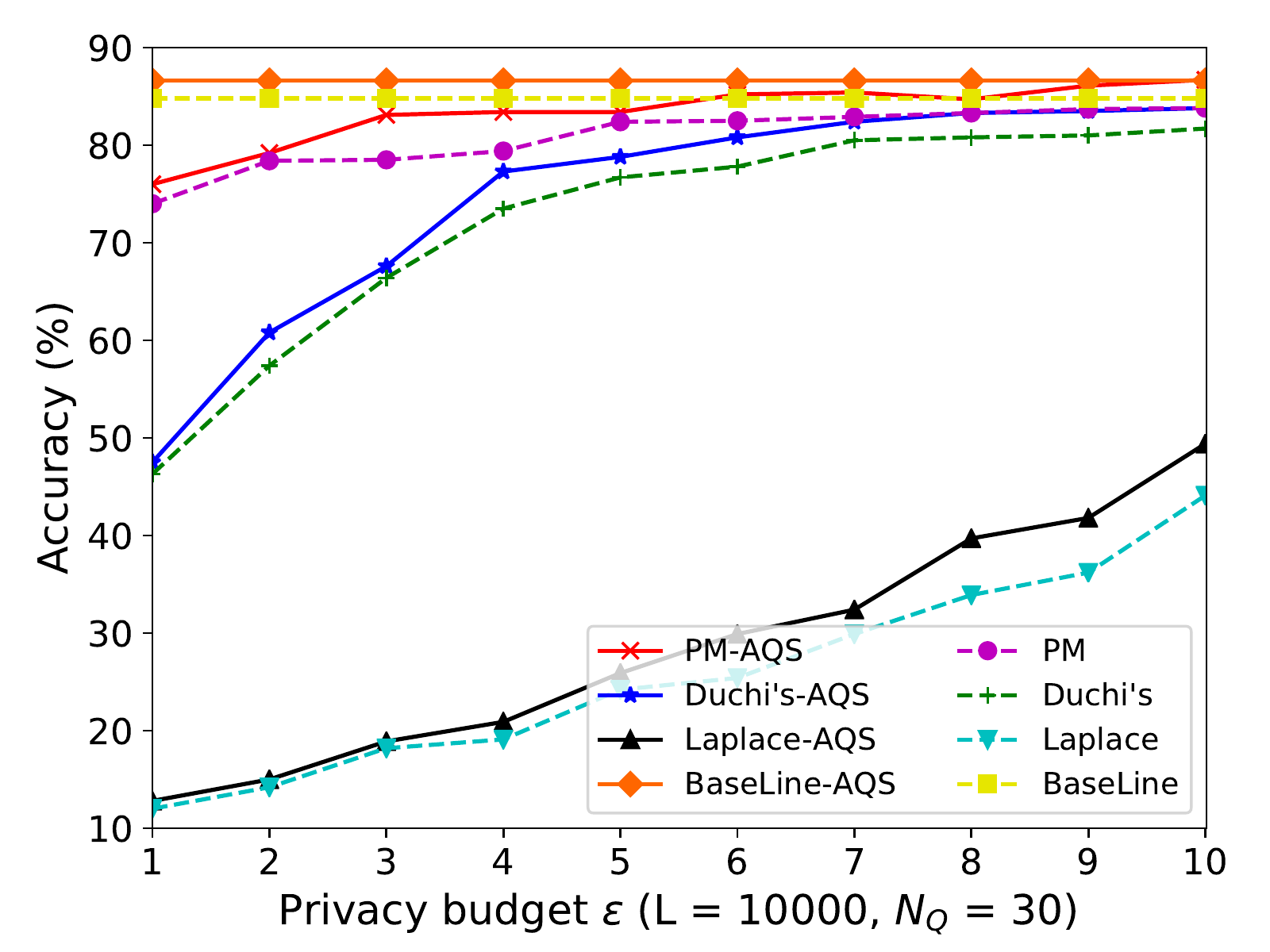}
		\caption{}
        \label{FashionMNIST_TotalBudget_smooth}
	\end{subfigure}
%%%%%%%%%%%%%%
	\begin{subfigure}{.3\textwidth}
		\includegraphics[width=\textwidth]{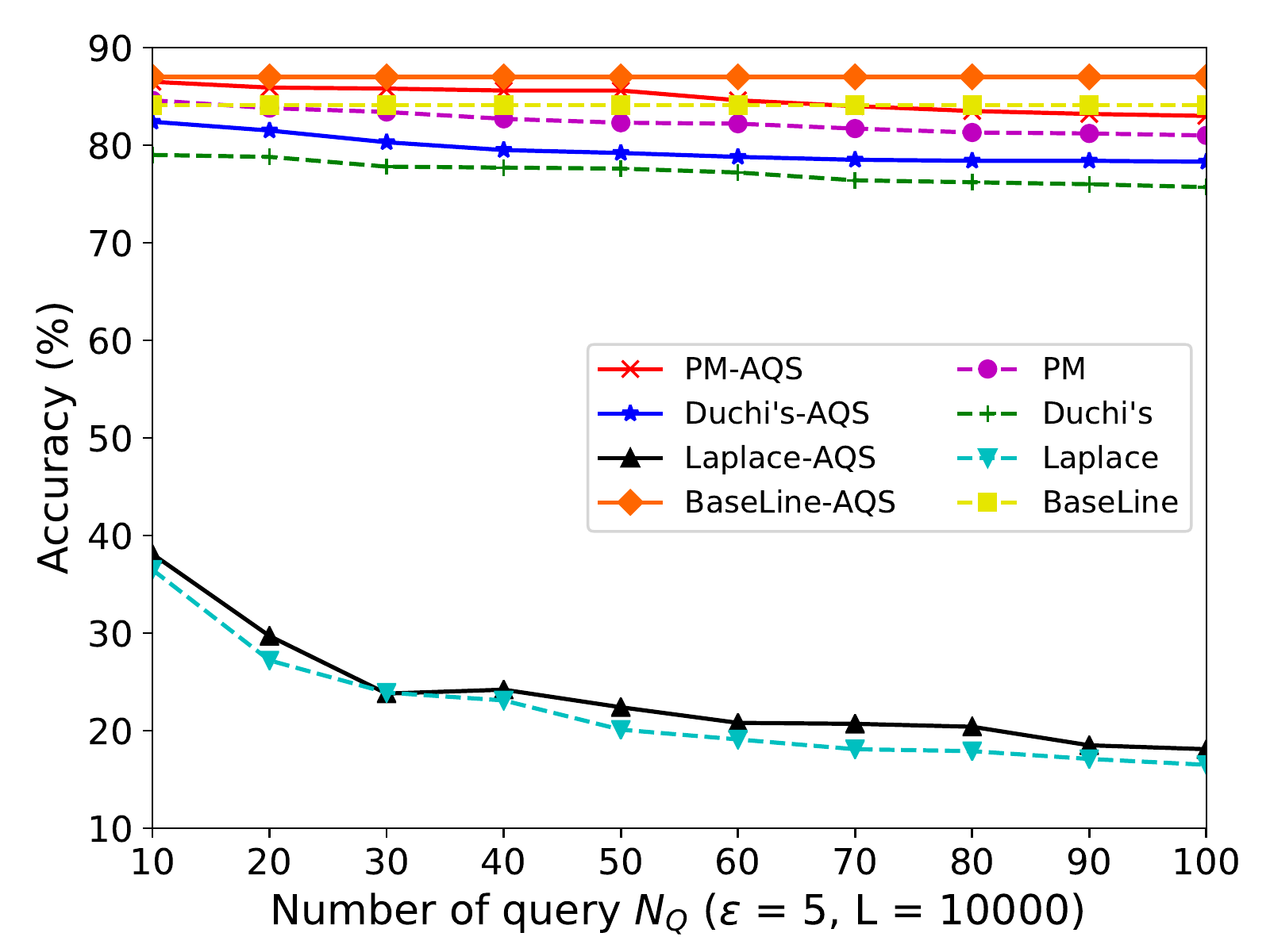}
		\caption{}
        \label{FashionMNIST_NumberofQuery_smooth}
	\end{subfigure}
%%%%%%%%%%%%%%
	\begin{subfigure}{.3\textwidth}
		\includegraphics[width=\textwidth]{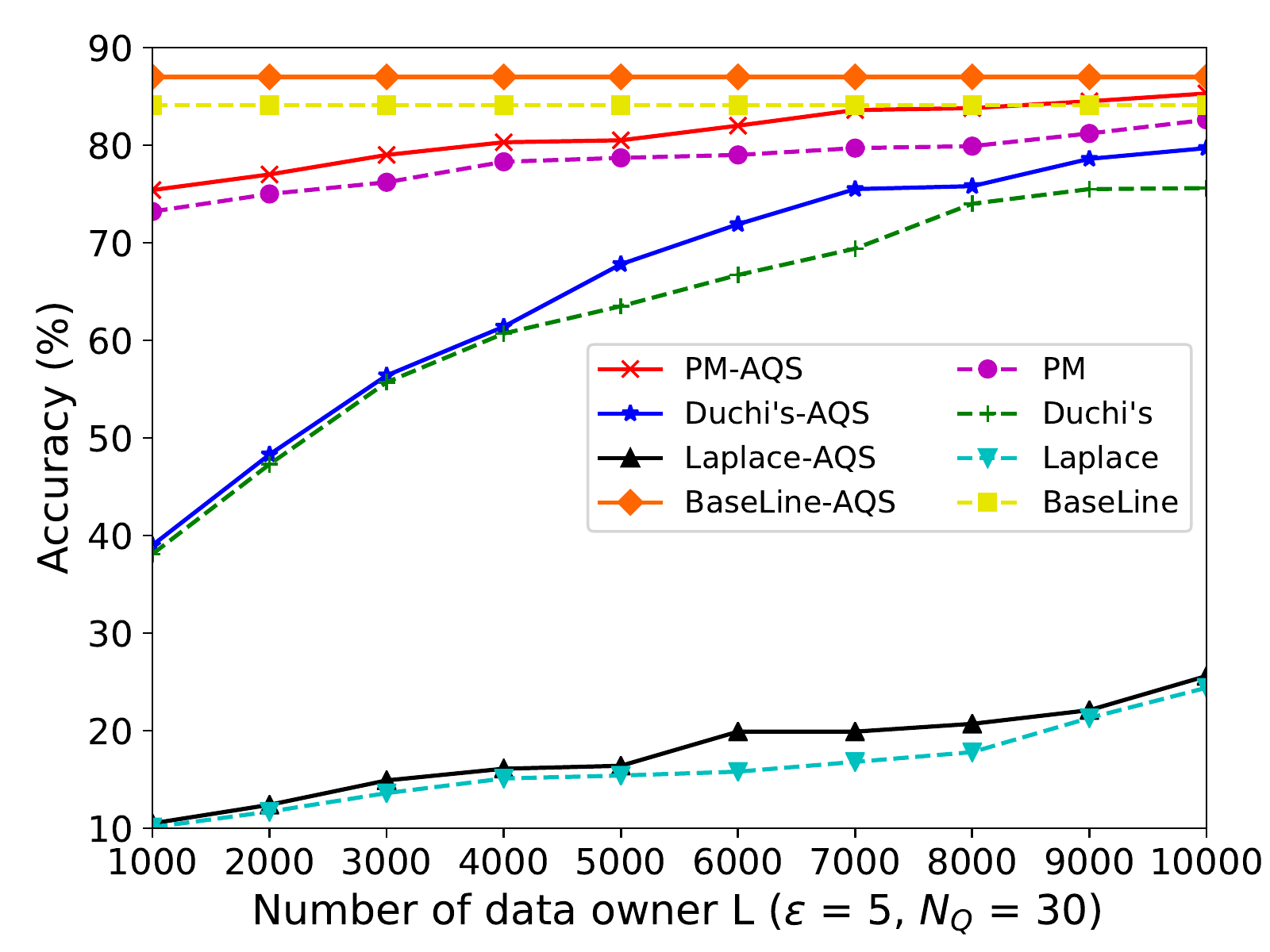}
		\caption{}
        \label{FashionMNIST_DataOwner_smooth}
	\end{subfigure}
%%%%%%%%%%%%%%
	\caption{LDP-DL experimental results on FashionMNIST dataset.}
    \label{FashionMNIST_dataset}
\end{figure*}

\subsection{Effectiveness Analysis}
In this section, we evaluate the effectiveness of our approach on three benchmark image datasets with different parameters and basic LDP mechanisms (as shown in Fig. \ref{CIFAR10_dataset}, Fig. \ref{MNIST_dataset} and Fig. \ref{FashionMNIST_dataset}). 
%Fig. \ref{CIFAR10_TotalBudget_smooth}, Fig. \ref{MNIST_TotalBudget_smooth} and Fig. \ref{FashionMNIST_TotalBudget_smooth} show the impact of the privacy budget on the results of our approach. As the privacy budget $\epsilon$ increasing, the accuracy increases, since less noise would be added to the data owners' perturbed distilled knowledge.  
%Fig. \ref{CIFAR10_NumberofQuery_smooth}, Fig. \ref{MNIST_NumberofQuery_smooth} and Fig. \ref{FashionMNIST_NumberofQuery_smooth} illustrate the results under different number of queries of each sample. As the number of queries increasing, the accuracy decreases, which is inline with our analysis in Section~\ref{sec:Privacy_Budget_Analysis}. 
%Fig. \ref{CIFAR10_DataOwner_smooth}, Fig. \ref{MNIST_DataOwner_smooth} and Fig. \ref{FashionMNIST_DataOwner_smooth} show the influence of the number of participated data owners on the performance of our approach. As more data owners participating in our framework, the accuracy tends to increase. 
We can observe that Piecewise mechanism always performs better than Duchi's mechanism and Laplace mechanism in our framework. Moreover, the results with our AQS process consistently outperforms the ones without our AQS process, which demonstrates that our proposed AQS could dramatically save the privacy budget and prevent privacy budget exploding from happening in privacy-preserving distributed deep learning training.

% Fig. \ref{CIFAR10_dataset}, \ref{MNIST_dataset} and \ref{FashionMNIST_dataset} show the student model accuracy results of LDP-DL running on 3 datasets with 3 LDP mechanism, respectively.
% For active query sampling (AQS) and random sampling, 3 parameters from different ranges have been evaluated:
% \begin{itemize}
%   \item Total privacy budget($\epsilon$): 1 to 10 at intervals of 1
%   \item Number of query(N): 10 to 100 at intervals of 10
%   \item Number of data owner(L): 1000 to 10000 at intervals of 1000
% \end{itemize}
%We observe that the active query sampling consistently outperforms the random sampling while different parameters variate. In 3 LDP mechanism, PM mechanism is the most effective mechanism for private query from teacher models.

\subsubsection{Effectiveness Analysis of Different Parameters}
The performance of our proposed LDP-DL framework is affected by multiple parameters ($\epsilon$, $N_{Q}$ and $L$). To evaluate the effectiveness of single parameter, as shown in Fig. \ref{CIFAR10_dataset}, Fig. \ref{MNIST_dataset} and Fig. \ref{FashionMNIST_dataset}, the other parameter are set as constant values. From Fig. \ref{CIFAR10_dataset}, Fig. \ref{MNIST_dataset} and Fig. \ref{FashionMNIST_dataset}, we observe that:

\begin{enumerate}
    \item The total privacy budget ($\epsilon$): This parameter controls %constraint limits 
    the noise scale of the private queries from each data owner's teacher model. We investigate $\epsilon$ from 1 to 10. Fig. \ref{CIFAR10_TotalBudget_smooth}, Fig. \ref{MNIST_TotalBudget_smooth} and Fig. \ref{FashionMNIST_TotalBudget_smooth} show the impact of the privacy budget on the results of our approach. As the privacy budget $\epsilon$ increasing, the accuracy increases, since less noise would be added to the data owners' perturbed distilled knowledge.
    %results in Fig. \ref{CIFAR10_dataset} (a), \ref{MNIST_dataset} (a) and \ref{FashionMNIST_dataset} (a) show the performance of student model while the privacy budget increasing. 
    In LDP mechanisms, greater $\epsilon$ results in smaller-scaled noise, and vice versa. While querying from multiple data owners' teacher models, the aggregation of the query results with smaller-scaled noise gives more information towards the data user's student model. Namely, the aggregated value is more close to the actual value, which benefits the training of the data user's student model. As such, a greater $\epsilon$ would result in a better accuracy performance.
    
    \item The number of queries ($N_{Q}$): This parameter controls the number of data owners' teacher models to be queried for each unlabelled public data. 
    Fig. \ref{CIFAR10_NumberofQuery_smooth}, Fig. \ref{MNIST_NumberofQuery_smooth} and Fig. \ref{FashionMNIST_NumberofQuery_smooth} illustrate the results under different number of queries of each public data. As the number of queries increasing, the accuracy decreases, which is inline with our analysis in Section~\ref{sec:Privacy_Budget_Analysis}. Because increasing the number of queries of each public data will actually result in more noise being added to each query result while maintaining the same total privacy budget. 
    % consume more privacy budget on each data owner
    % Fig. \ref{CIFAR10_dataset} (b), \ref{MNIST_dataset} (b) and \ref{FashionMNIST_dataset} (b) plot performance of student model while the number of query ascent from 10 to 100. While fixing the total privacy budget and number of data owner, increasing the number of query will consume more privacy budget on each data owner. 
    Specifically speaking, as shown in our results, as $N_{Q}$ increasing, the accuracy of the data user's student model only slightly declines while utilizing either Piecewise mechanism or Duchi's mechanism. However, for the Laplace mechanism, the accuracy significantly decreases while $N_{Q}$ increasing. Because different LDP mechanisms would result in different noise scale in terms of $N_{Q}$. Compared with Laplace mechanism, Using Piecewise mechanism and Duchi's mechanism would decrease the influence of $N_{Q}$ on the performance of our LDP-DL framework.
    
    % This caused by the noise scale introduce by different LDP mechanism. With a relative large number of data owner and suitable LDP mechanism, the effect of the number of query can be mitigated.
    
    \item The number of data owners ($L$): This parameter indicates the number of data owners participated in our LDP-DL framework. 
    As analyzed in section \ref{sec:Privacy_Budget_Analysis}, increasing the number of participated data owners can reduce the overall noise of the aggregated information. 
    Fig. \ref{CIFAR10_DataOwner_smooth}, Fig. \ref{MNIST_DataOwner_smooth} and Fig. \ref{FashionMNIST_DataOwner_smooth} show the influence of the number of participated data owners on the performance of our approach. As more data owners participating in our framework, the accuracy tends to increase. 
    %Fig. \ref{CIFAR10_dataset} (c), \ref{MNIST_dataset} (c) and \ref{FashionMNIST_dataset} (c) show the performance of student model while the number of data owner increases from 1000 to 10000. 
    %As analyzed in section \ref{sec:Privacy_Budget_Analysis}, increasing the number of participated data owners can reduce the overall noise of the aggregated information. It can be observed that the accuracy of the student model increases while the number of data owner raises. 
    Since the total privacy budget is controlled by each data owner's preference, to obtain an appropriate performance of the data user's student model, the number of participated data owners of our LDP-DL framework should be set to a sufficient value.
\end{enumerate}

% \subsection{Efficiency Analysis}
% In LDP-DL framework, we assume the teachers already have their models trained on the corresponding private data. The student is trained with active query sampling on public data with labels that come from the aggregation of teachers. Therefore, the main time consuming part is the interaction between student (data user) and teachers (data owners). Assuming the batch size of active query sampling is $K$, the total number of public data is $N$ and $M$ teachers have been queried for each public data. There are two setting can be applied to active query sampling:
% \begin{itemize}
%     \item Fix the iteration of training: Given a fix iteration of active query sampling $T$, the total communications is fixed, the total communications between student and teachers is $T*K*M$ and the time complexity is $O(TKM)$. 
%     \item Fix the accuracy baseline of training: If the accuracy requirement is provided, the worst case happens when there are no more public data available for further training. In this case, the total communications   between student and teachers is $N*M$ and time complexity is $O(NM)$.   
% \end{itemize}

\begin{table*}[h]
    \centering
    \begin{tabular}{c|c|c|c|c|c|c}
        \hline
        Datasets & \multicolumn{2}{|c}{CIFAR10~\cite{krizhevsky2009learning}} & \multicolumn{2}{|c}{MNIST~\cite{lecun1998gradient}} & \multicolumn{2}{|c}{FashionMNIST~\cite{xiao2017fashion}}\\
        \hline
        Approaches & Accuracy & Privacy Budget & Accuracy & Privacy Budget & Accuracy & Privacy Budget \\
        \hline\hline
        \multirow{2}{*}{LDP-DL} & 77.5\% & 5 & 98.1\% & 5 & 83.4\% & 5 \\
        \cline{2-7}
         & 79.7\% & 8 & 98.8\% & 8 & 85.7\% & 8 \\
        \hline
        DP-SGD~\cite{abadi2016deep} & 73.0\% & 8 & 97.00\% & 8 & - & -  \\
        \hline
        \multirow{2}{*}{PATE~\cite{papernot2016semi}} & 73.6\% & 5 & 97.7\% & 5 & 81.5\% & 5  \\
        \cline{2-7}
        & 76.0\% & 8 & 98.2\% & 8 & 84.7\% & 8  \\
        \hline
        % PATE\_Scaled & - & - & 98.5\% & 1.97 & - & -  \\
        % \hline
        \multirow{2}{*}{DP-FL~\cite{geyer2017differentially}} & 75.9\% & 5 & 96.4\% & 5 & 82.6\% & 5  \\
        \cline{2-7}
        & 78.7\% & 8 & 97.2\% & 8 & 83.6\% & 8  \\
         \hline
        % PET-DL & 76.81\% & 5.48 & 99.3\% & 1.21 & - & -  \\
        % \hline
        % RONA & 79.0\% & 5 & 98.2\% & 5 & - & -  \\
        
    \end{tabular}
    \caption{In Comparison with Existing Approaches.}
    \label{Comparison_SOTA}
\end{table*}

\subsection{In Comparison with Existing Approaches}
In this section, we have compared our LDP-DL framework with 3 state-of-the-art approaches: DP-SGD \cite{abadi2016deep}, PATE \cite{papernot2016semi} and DP-FL \cite{geyer2017differentially}. 
\begin{itemize}
  \item DP-SGD~\cite{abadi2016deep}: Differentially Private Stochastic Gradient Descent (DP-SGD) algorithm trains the deep neural network with differential privacy under a centralized setting. It utilizes the Gaussian mechanism on random subset of examples to produce average noisy gradient for model optimization. This approach does not have available code from public resource. Therefore, we directly refer the results presented in the original paper.
  \item PATE~\cite{papernot2016semi}: Private Aggregation of Teacher Ensembles (PATE) proposed a distributed teacher-student framework. The privacy guarantee comes from the perturbation on teachers voting aggregation. The ensemble decision based on the noisy voting provides the label of student model's training data. The student model is trained on semi-supervised learning with GANs. PATE~\cite{papernot2016semi} are evaluated on the code provided by the paper authors.
  \item DP-FL~\cite{geyer2017differentially}: Differentially Private Federated Learning (DL-FL) describes a federated optimization algorithm under private manner. Instead of directly averaging the distributed client models updates, an alter approach that use random sampling and Gaussian mechanism on sum of clients updates is introduced to approximate the averaging. The curator collects the noisy updates to optimize the center server model. Since the original paper aims at protecting the privacy at the client's level but not at the sample's level, DP-FL~\cite{geyer2017differentially} are evaluated on the code published by the author with some minor changes to enable privacy preservation at the sample's level. % We apply the same distorting on the singe updates to protect the single data point of client's dataset.
\end{itemize}

Table \ref{Comparison_SOTA} shows the results of different approaches under the same privacy budgets. 
The results of LDP-DL are evaluated under parameters $L=10,000$, $N_{Q}=30$. For the other approaches, we strictly follow the settings mentioned in the corresponding papers and keep the common parameters (such as the number of data owners (clients), privacy budgets) at the same level. Under the same level of privacy budgets, we can observe that LDP-DL consistently outperforms the other competitors. The improvement can be derived from the knowledge distillation and active query sampling. Knowledge distillation leverage richer information while transferring the knowledge from the teacher models to a student model. Meanwhile, active query sampling efficiently reduces the total number of queries from the data user (i.e., the student model) to the data owners (i.e., the teacher models). As such, the total cost of privacy budget is been reduced dramatically. 
% namely, the overall noise has been reduced. 

\section{Related Work} \label{sec:ldp_relatedWork}
Local Differential Privacy (LDP) has been proposed \cite{duchi2013local} to remove the trusted curator of the centralized differential privacy. LDP also provides the data owner more controls on the information left hands in a more strict and realistic privacy manner. LDP for statistical information collection and estimation have been well studied in the past decades \cite{warner1965randomized, kairouz2014extremal, wang2017locally, bassily2015local, erlingsson2014rappor, cormode2018privacy, ding2017collecting, bittau2017prochlo}. 

Recently, more works propose to apply DP or LDP in data mining and machine learning applications, such as clustering \cite{nissim2018clustering}, Bayesian inference \cite{yilmaz2019locally}, frequent itemset mining \cite{wang2018locally} and probability distribution estimation \cite{kairouz2016discrete, murakami2018toward, ye2018optimal, wang2018locally}. However, only a few recent works aim to use LDP in deep learning. 
For instance, Abadi et al. \cite{abadi2016deep} propose to train the deep neural network via stochastic gradient descent with differential privacy under a centralized setting. However, it not only requires an impractical trusted third party to serve as the trusted curator but also has the privacy budget exploding issue (i.e., causing impractical huge privacy budget to train a meaningful deep learning model).

Papernot et al. \cite{papernot2016semi} proposes PATE, a ``teacher-student'' paradigm for privacy-preserving deep learning, where each data owner learns a teacher model using its own (local) private dataset, and the data user aims to learn a student model using the unlabelled public data (but no direct access to the data owners' private data) to mimic the output of the ensemble of the teacher models, i.e., the student learns to make predictions that is the same as the most number of teachers. To ensure privacy, PATE \cite{papernot2016semi} assumes a trusted aggregator to provide a differentially private query interface, where the data user could query the ensemble of the teacher models (from the data owners) using the unlabelled public data to obtain the labels for the training of the student model. However, a fully trusted aggregator barely exists in most of the real-world distributed deep learning scenarios. 
Chase et al. \cite{chase2017private} proposes a private collaborative neural network learning approach, that combines secure multi-party computation (MPC), differential privacy (DP) and secret sharing. Since the MPC protocol is implemented via a garbled circuit whose size is subject to the number of parameters (i.e., the size of the gradient) of the neural network, it tends to be less efficient and not scalable while training larger neural networks. Also, in \cite{chase2017private}, using secret sharing requires at least two non-colluding honest data users which might not be practical. 
In \cite{geyer2017differentially}, the authors present a federated optimization algorithm under private manner. Instead of directly averaging the distributed client models updates, an alter approach that use random sampling and Gaussian mechanism on sum of clients updates is introduced to approximate the averaging. The curator collects the noisy updates to optimize the center server model. However, this approach only focus on training small deep learning models (i.e., only training one or very few number of iterations) and easier datasets (i.e., not testing on any image datasets).

Our approach aims to solve the challenges left by the previous approaches. The difference could be summarized in three folds: (i) our approach aims to enable training large deep neural networks (e.g., ResNet) on popular benchmark image datasets (e.g., CIFAR-10); (ii) our approach designs a proactive mechanism (i.e., the active query sampling) to reduce the overall privacy budget efficiently to prevent privacy budget exploding while training large deep neural networks; (iii) our approach is not based on federated learning, thus does not have to satisfy the requirements of performing federated learning (e.g., clients being online around the same time period).

\section{Conclusion} \label{sec:ldp_conclusion}
In this paper, we proposed LDP-DL, a novel, effective and efficient privacy-preserving distributed deep learning framework using local differential privacy and knowledge distillation. We also present an active sampling approach to efficiently reduce the total number of queries from the data user to each data owners, so that to reduce the total cost of privacy budget. In the experimental evaluation, a comprehensive comparison has been made among our algorithm and three state-of-the-art privacy-preserving deep learning approaches. Extensive experiments have been conducted on three benchmark image datasets. Our results show that LDP-DL consistently outperforms the other competitors in terms of privacy budget and model accuracy.

% or


%\appendices
%% you can choose not to have a title for an appendix
%% if you want by leaving the argument blank
%\section{}
%% use section* for acknowledgment
%\ifCLASSOPTIONcompsoc
%  % The Computer Society usually uses the plural form
%\else
%  % regular IEEE prefers the singular form
%\fi

\ifCLASSOPTIONcaptionsoff
  \newpage
\fi



%\bibliographystyle{IEEEtran}
%\begin{thebibliography}{1}
%\end{thebibliography}

 \bibliographystyle{IEEEtran}
 \bibliography{IEEEabrv,ldp}

\begin{thebibliography}{10}
\providecommand{\url}[1]{#1}
\csname url@samestyle\endcsname
\providecommand{\newblock}{\relax}
\providecommand{\bibinfo}[2]{#2}
\providecommand{\BIBentrySTDinterwordspacing}{\spaceskip=0pt\relax}
\providecommand{\BIBentryALTinterwordstretchfactor}{4}
\providecommand{\BIBentryALTinterwordspacing}{\spaceskip=\fontdimen2\font plus
\BIBentryALTinterwordstretchfactor\fontdimen3\font minus
  \fontdimen4\font\relax}
\providecommand{\BIBforeignlanguage}[2]{{%
\expandafter\ifx\csname l@#1\endcsname\relax
\typeout{** WARNING: IEEEtran.bst: No hyphenation pattern has been}%
\typeout{** loaded for the language `#1'. Using the pattern for}%
\typeout{** the default language instead.}%
\else
\language=\csname l@#1\endcsname
\fi
#2}}
\providecommand{\BIBdecl}{\relax}
\BIBdecl

\bibitem{perez2019solo}
F.~Perez, S.~Avila, and E.~Valle, ``Solo or ensemble? choosing a cnn
  architecture for melanoma classification,'' in \emph{Proceedings of the IEEE
  Conference on Computer Vision and Pattern Recognition Workshops}, 2019, pp.
  0--0.

\bibitem{wu2016cost}
P.-Y. Wu, C.-C. Fang, J.~M. Chang, and S.-Y. Kung, ``Cost-effective kernel
  ridge regression implementation for keystroke-based active authentication
  system,'' \emph{IEEE transactions on cybernetics}, vol.~47, no.~11, pp.
  3916--3927, 2016.

\bibitem{nguyen2019autogan}
H.~Nguyen, D.~Zhuang, P.-Y. Wu, and M.~Chang, ``Autogan-based dimension
  reduction for privacy preservation,'' \emph{Neurocomputing}, 2019.

\bibitem{zhuang2017fripal}
D.~Zhuang, S.~Wang, and J.~M. Chang, ``Fripal: Face recognition in privacy
  abstraction layer,'' in \emph{2017 IEEE Conference on Dependable and Secure
  Computing}.\hskip 1em plus 0.5em minus 0.4em\relax IEEE, 2017, pp. 441--448.

\bibitem{zhuang2017peerhunter}
D.~Zhuang and J.~M. Chang, ``Peerhunter: Detecting peer-to-peer botnets through
  community behavior analysis,'' in \emph{2017 IEEE Conference on Dependable
  and Secure Computing}.\hskip 1em plus 0.5em minus 0.4em\relax IEEE, 2017, pp.
  493--500.

\bibitem{zhuang2018enhanced}
------, ``Enhanced peerhunter: Detecting peer-to-peer botnets through
  network-flow level community behavior analysis,'' \emph{IEEE Transactions on
  Information Forensics and Security}, vol.~14, no.~6, pp. 1485--1500, 2018.

\bibitem{zhuang2019dynamo}
D.~Zhuang, M.~J. Chang, and M.~Li, ``Dynamo: Dynamic community detection by
  incrementally maximizing modularity,'' \emph{IEEE Transactions on Knowledge
  and Data Engineering}, 2019.

\bibitem{8835245}
M.~{Nasr}, R.~{Shokri}, and A.~{Houmansadr}, ``Comprehensive privacy analysis
  of deep learning: Passive and active white-box inference attacks against
  centralized and federated learning,'' in \emph{2019 IEEE Symposium on
  Security and Privacy (SP)}, May 2019, pp. 739--753.

\bibitem{shokri2017membership}
R.~Shokri, M.~Stronati, C.~Song, and V.~Shmatikov, ``Membership inference
  attacks against machine learning models,'' in \emph{2017 IEEE Symposium on
  Security and Privacy (SP)}.\hskip 1em plus 0.5em minus 0.4em\relax IEEE,
  2017, pp. 3--18.

\bibitem{papernot2016semi}
N.~Papernot, M.~Abadi, U.~Erlingsson, I.~Goodfellow, and K.~Talwar,
  ``Semi-supervised knowledge transfer for deep learning from private training
  data,'' \emph{arXiv preprint arXiv:1610.05755}, 2016.

\bibitem{chase2017private}
M.~Chase, R.~Gilad-Bachrach, K.~Laine, K.~E. Lauter, and P.~Rindal, ``Private
  collaborative neural network learning.'' \emph{IACR Cryptology ePrint
  Archive}, vol. 2017, p. 762, 2017.

\bibitem{geyer2017differentially}
R.~C. Geyer, T.~Klein, and M.~Nabi, ``Differentially private federated
  learning: A client level perspective,'' \emph{arXiv preprint
  arXiv:1712.07557}, 2017.

\bibitem{wang2019collecting}
N.~Wang, X.~Xiao, Y.~Yang, J.~Zhao, S.~C. Hui, H.~Shin, J.~Shin, and G.~Yu,
  ``Collecting and analyzing multidimensional data with local differential
  privacy,'' in \emph{2019 IEEE 35th International Conference on Data
  Engineering (ICDE)}.\hskip 1em plus 0.5em minus 0.4em\relax IEEE, 2019, pp.
  638--649.

\bibitem{hinton2015distilling}
G.~Hinton, O.~Vinyals, and J.~Dean, ``Distilling the knowledge in a neural
  network,'' \emph{arXiv preprint arXiv:1503.02531}, 2015.

\bibitem{abadi2016deep}
M.~Abadi, A.~Chu, I.~Goodfellow, H.~B. McMahan, I.~Mironov, K.~Talwar, and
  L.~Zhang, ``Deep learning with differential privacy,'' in \emph{Proceedings
  of the 2016 ACM SIGSAC Conference on Computer and Communications Security},
  2016, pp. 308--318.

\bibitem{krizhevsky2009learning}
A.~Krizhevsky, G.~Hinton \emph{et~al.}, ``Learning multiple layers of features
  from tiny images,'' 2009.

\bibitem{lecun1998gradient}
Y.~LeCun, L.~Bottou, Y.~Bengio, and P.~Haffner, ``Gradient-based learning
  applied to document recognition,'' \emph{Proceedings of the IEEE}, vol.~86,
  no.~11, pp. 2278--2324, 1998.

\bibitem{xiao2017fashion}
H.~Xiao, K.~Rasul, and R.~Vollgraf, ``Fashion-mnist: a novel image dataset for
  benchmarking machine learning algorithms,'' \emph{arXiv preprint
  arXiv:1708.07747}, 2017.

\bibitem{cynthia2006differential}
D.~Cynthia, ``Differential privacy,'' \emph{Automata, languages and
  programming}, pp. 1--12, 2006.

\bibitem{dwork2014algorithmic}
C.~Dwork, A.~Roth \emph{et~al.}, ``The algorithmic foundations of differential
  privacy,'' \emph{Foundations and Trends{\textregistered} in Theoretical
  Computer Science}, vol.~9, no. 3--4, pp. 211--407, 2014.

\bibitem{dwork2014analyze}
C.~Dwork, K.~Talwar, A.~Thakurta, and L.~Zhang, ``Analyze gauss: optimal bounds
  for privacy-preserving principal component analysis,'' in \emph{Proceedings
  of the forty-sixth annual ACM symposium on Theory of computing}, 2014, pp.
  11--20.

\bibitem{duchi2013local}
J.~C. Duchi, M.~I. Jordan, and M.~J. Wainwright, ``Local privacy and
  statistical minimax rates,'' in \emph{2013 IEEE 54th Annual Symposium on
  Foundations of Computer Science}.\hskip 1em plus 0.5em minus 0.4em\relax
  IEEE, 2013, pp. 429--438.

\bibitem{wang2017locally}
T.~Wang, J.~Blocki, N.~Li, and S.~Jha, ``Locally differentially private
  protocols for frequency estimation,'' in \emph{26th $\{$USENIX$\}$ Security
  Symposium ($\{$USENIX$\}$ Security 17)}, 2017, pp. 729--745.

\bibitem{ba2014deep}
J.~Ba and R.~Caruana, ``Do deep nets really need to be deep?'' in
  \emph{Advances in neural information processing systems}, 2014, pp.
  2654--2662.

\bibitem{polino2018model}
A.~Polino, R.~Pascanu, and D.~Alistarh, ``Model compression via distillation
  and quantization,'' \emph{arXiv preprint arXiv:1802.05668}, 2018.

\bibitem{biggio2012poisoning}
B.~Biggio, B.~Nelson, and P.~Laskov, ``Poisoning attacks against support vector
  machines,'' \emph{arXiv preprint arXiv:1206.6389}, 2012.

\bibitem{gu2017badnets}
T.~Gu, B.~Dolan-Gavitt, and S.~Garg, ``Badnets: Identifying vulnerabilities in
  the machine learning model supply chain,'' \emph{arXiv preprint
  arXiv:1708.06733}, 2017.

\bibitem{chen2017targeted}
X.~Chen, C.~Liu, B.~Li, K.~Lu, and D.~Song, ``Targeted backdoor attacks on deep
  learning systems using data poisoning,'' \emph{arXiv preprint
  arXiv:1712.05526}, 2017.

\bibitem{liu2017neural}
Y.~Liu, Y.~Xie, and A.~Srivastava, ``Neural trojans,'' in \emph{2017 IEEE
  International Conference on Computer Design (ICCD)}.\hskip 1em plus 0.5em
  minus 0.4em\relax IEEE, 2017, pp. 45--48.

\bibitem{deng2009imagenet}
J.~Deng, W.~Dong, R.~Socher, L.-J. Li, K.~Li, and L.~Fei-Fei, ``Imagenet: A
  large-scale hierarchical image database,'' in \emph{2009 IEEE conference on
  computer vision and pattern recognition}.\hskip 1em plus 0.5em minus
  0.4em\relax Ieee, 2009, pp. 248--255.

\bibitem{duchi2018minimax}
J.~C. Duchi, M.~I. Jordan, and M.~J. Wainwright, ``Minimax optimal procedures
  for locally private estimation,'' \emph{Journal of the American Statistical
  Association}, vol. 113, no. 521, pp. 182--201, 2018.

\bibitem{mcsherry2007mechanism}
F.~McSherry and K.~Talwar, ``Mechanism design via differential privacy,'' in
  \emph{48th Annual IEEE Symposium on Foundations of Computer Science
  (FOCS'07)}.\hskip 1em plus 0.5em minus 0.4em\relax IEEE, 2007, pp. 94--103.

\bibitem{settles2009active}
B.~Settles, ``Active learning literature survey,'' University of
  Wisconsin-Madison Department of Computer Sciences, Tech. Rep., 2009.

\bibitem{yang2012differential}
Y.~Yang, Z.~Zhang, G.~Miklau, M.~Winslett, and X.~Xiao, ``Differential privacy
  in data publication and analysis,'' in \emph{Proceedings of the 2012 ACM
  SIGMOD International Conference on Management of Data}, 2012, pp. 601--606.

\bibitem{warner1965randomized}
S.~L. Warner, ``Randomized response: A survey technique for eliminating evasive
  answer bias,'' \emph{Journal of the American Statistical Association},
  vol.~60, no. 309, pp. 63--69, 1965.

\bibitem{kairouz2014extremal}
P.~Kairouz, S.~Oh, and P.~Viswanath, ``Extremal mechanisms for local
  differential privacy,'' \emph{arXiv preprint arXiv:1407.1338}, 2014.

\bibitem{bassily2015local}
R.~Bassily and A.~Smith, ``Local, private, efficient protocols for succinct
  histograms,'' in \emph{Proceedings of the forty-seventh annual ACM symposium
  on Theory of computing}, 2015, pp. 127--135.

\bibitem{erlingsson2014rappor}
{\'U}.~Erlingsson, V.~Pihur, and A.~Korolova, ``Rappor: Randomized aggregatable
  privacy-preserving ordinal response,'' in \emph{Proceedings of the 2014 ACM
  SIGSAC conference on computer and communications security}, 2014, pp.
  1054--1067.

\bibitem{cormode2018privacy}
G.~Cormode, S.~Jha, T.~Kulkarni, N.~Li, D.~Srivastava, and T.~Wang, ``Privacy
  at scale: Local differential privacy in practice,'' in \emph{Proceedings of
  the 2018 International Conference on Management of Data}, 2018, pp.
  1655--1658.

\bibitem{ding2017collecting}
B.~Ding, J.~Kulkarni, and S.~Yekhanin, ``Collecting telemetry data privately,''
  \emph{arXiv preprint arXiv:1712.01524}, 2017.

\bibitem{bittau2017prochlo}
A.~Bittau, {\'U}.~Erlingsson, P.~Maniatis, I.~Mironov, A.~Raghunathan, D.~Lie,
  M.~Rudominer, U.~Kode, J.~Tinnes, and B.~Seefeld, ``Prochlo: Strong privacy
  for analytics in the crowd,'' in \emph{Proceedings of the 26th Symposium on
  Operating Systems Principles}, 2017, pp. 441--459.

\bibitem{nissim2018clustering}
K.~Nissim and U.~Stemmer, ``Clustering algorithms for the centralized and local
  models,'' in \emph{Algorithmic Learning Theory}.\hskip 1em plus 0.5em minus
  0.4em\relax PMLR, 2018, pp. 619--653.

\bibitem{yilmaz2019locally}
E.~Yilmaz, M.~Al-Rubaie, and J.~M. Chang, ``Locally differentially private
  naive bayes classification,'' \emph{arXiv preprint arXiv:1905.01039}, 2019.

\bibitem{wang2018locally}
T.~Wang, N.~Li, and S.~Jha, ``Locally differentially private frequent itemset
  mining,'' in \emph{2018 IEEE Symposium on Security and Privacy (SP)}.\hskip
  1em plus 0.5em minus 0.4em\relax IEEE, 2018, pp. 127--143.

\bibitem{kairouz2016discrete}
P.~Kairouz, K.~Bonawitz, and D.~Ramage, ``Discrete distribution estimation
  under local privacy,'' in \emph{International Conference on Machine
  Learning}.\hskip 1em plus 0.5em minus 0.4em\relax PMLR, 2016, pp. 2436--2444.

\bibitem{murakami2018toward}
T.~Murakami, H.~Hino, and J.~Sakuma, ``Toward distribution estimation under
  local differential privacy with small samples,'' \emph{Proceedings on Privacy
  Enhancing Technologies}, vol. 2018, no.~3, pp. 84--104, 2018.

\bibitem{ye2018optimal}
M.~Ye and A.~Barg, ``Optimal schemes for discrete distribution estimation under
  locally differential privacy,'' \emph{IEEE Transactions on Information
  Theory}, vol.~64, no.~8, pp. 5662--5676, 2018.

\end{thebibliography}
\begin{IEEEbiography}[{\includegraphics[width=1.05in,height=2.0in,clip,keepaspectratio]{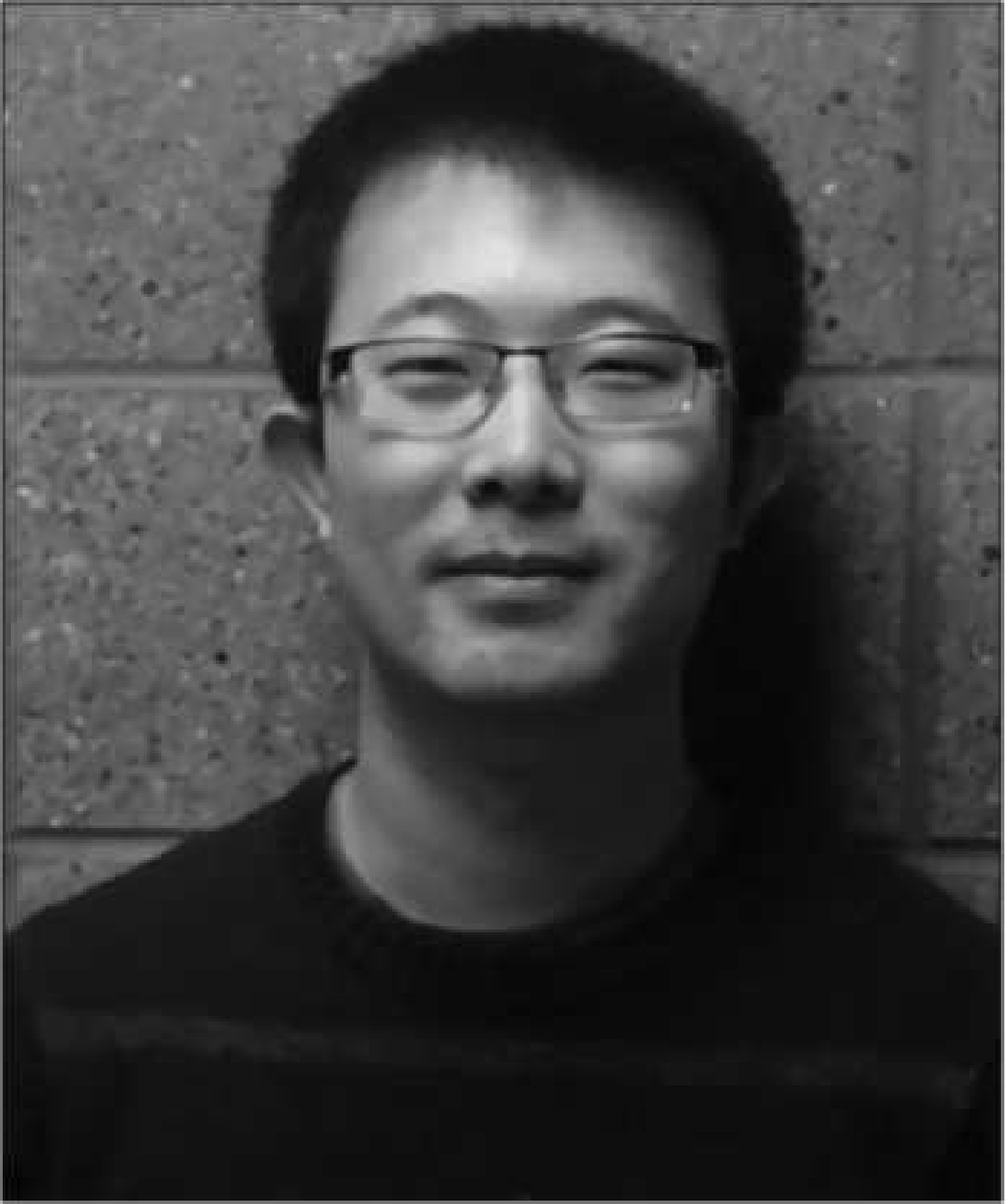}}]{Di Zhuang}
(S'15) is currently a Security and Privacy Engineer at Snap Inc. He received his Ph.D. degree in electrical engineering, and B.E. degree in computer science and information security from University of South Florida, Tampa and Nankai University, Tianjin, China, respectively. His research interests include network security, social network science, and privacy preserving machine learning. He is a member of IEEE. 
\end{IEEEbiography}
\vskip 0pt plus -1fil
\begin{IEEEbiography}[{\includegraphics[width=1.05in,height=2.0in,clip,keepaspectratio]{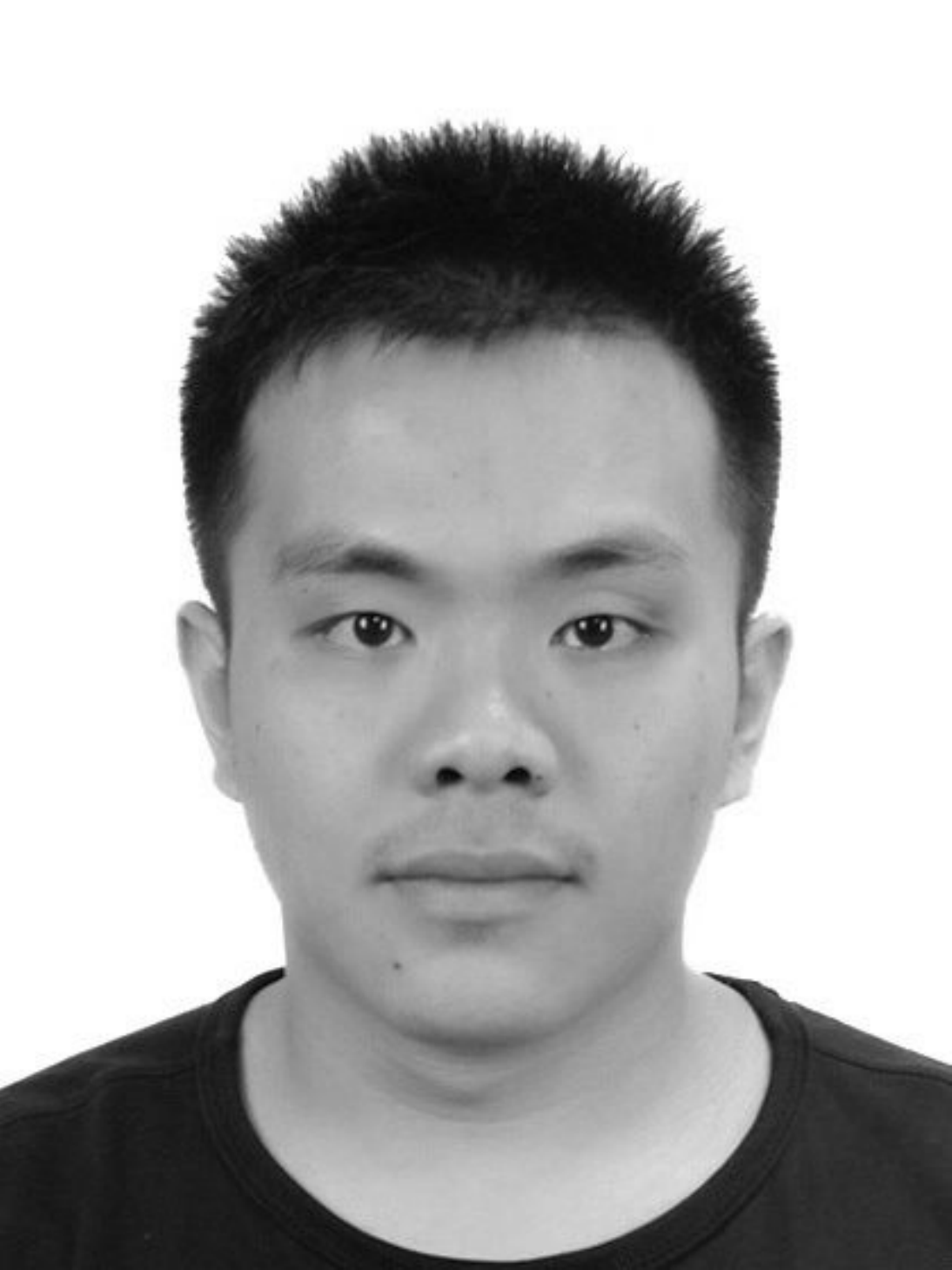}}]{Mingchen Li} received his M.S. degree in electrical engineering from Illinois Institute of Technology. He is currently pursuing the Ph.D. degree in electrical engineering with University of South Florida, Tampa. His research interests include cyber security, synthetic data generation, privacy enhancing technologies, machine learning and data analytics.
\end{IEEEbiography}
\vskip 0pt plus -1fil
\begin{IEEEbiography}[{\includegraphics[width=1.05in,height=2.0in,clip,keepaspectratio]{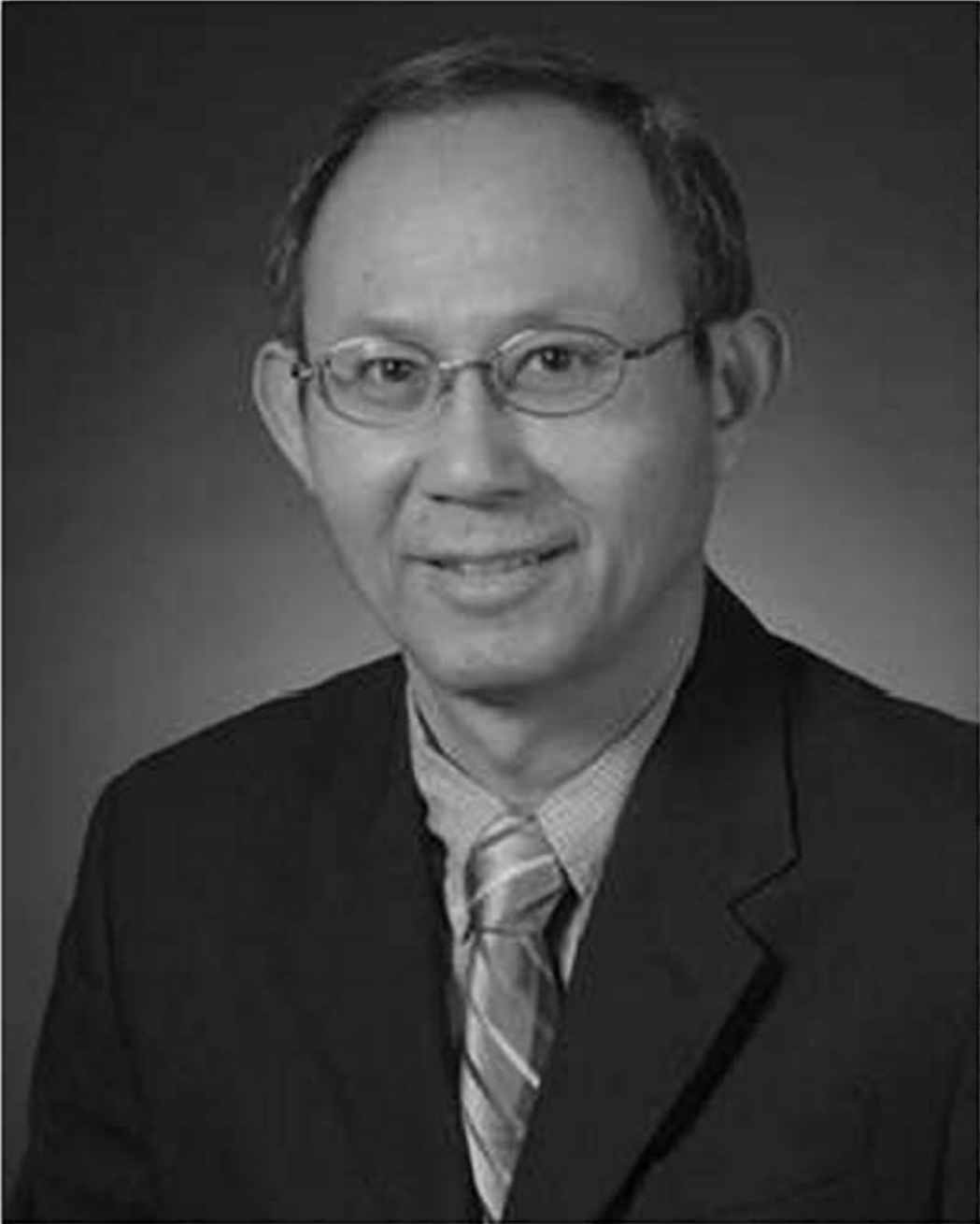}}]{J. Morris Chang}
(SM'08) is a professor in the Department of Electrical Engineering at the University of South Florida. He received the Ph.D. degree from the North Carolina State University. His past industrial experiences include positions at Texas Instruments, Microelectronic Center of North Carolina and AT\&T Bell Labs. He received the University Excellence in Teaching Award at Illinois Institute of Technology in 1999. His research interests include: cyber security, wireless networks, and energy efficient computer systems. In the last six years, his research projects on cyber security have been funded by DARPA. Currently, he is leading a DARPA project under Brandeis program focusing on privacy-preserving computation over Internet. He is a handling editor of Journal of Microprocessors and Microsystems and an editor of IEEE IT Professional. He is a senior member of IEEE.
\end{IEEEbiography}
\end{document}